\documentclass[natbib,twocolumn]{svjour3}  
\smartqed

\usepackage{booktabs}
\usepackage{graphicx}
\usepackage{amsmath}
\usepackage{amssymb}
\usepackage{dsfont}
\usepackage{setspace}
\usepackage{xspace}
 \usepackage{caption}
\usepackage{subcaption}
\usepackage{color}
\graphicspath{ {./images/} }

\newcommand*{\eg}{e.g.\@\xspace}
\newcommand*{\ie}{i.e.\@\xspace}

\DeclareMathOperator*{\argmax}{arg\,max}
\let\cite\citep

\journalname{IJCV}

\begin{document}
%\title{Unsupervised Grounding of Large-Video Collections}
\title{Unsupervised Semantic Action Discovery from Video Collections}
%  using Unsupervised Learning}


\author{Ozan~Sener \and Amir~Roshan~Zamir  \and Chenxia Wu \and Silvio~Savarese  \and Ashutosh~Saxena}
\authorrunning{Sener et al.}

\institute{
O.~Sener \at
Cornell University, Ithaca NY 14853, USA \\
\email{ozan@cs.cornell.edu}
\and
A.R.~Zamir \at
Stanford University, Stanford CA 94305, USA \\
\email{zamir@cs.stanford.edu}
\and
C.~Wu \at
Cornell University, Ithaca NY 14853, USA \\
\email{wu@cs.cornell.edu}
\and
S.~Savarese \at
Stanford University, Stanford CA 94305, USA  \\
\email{silvio@stanford.edu}
\and
A.~Saxena \at
Brain of Things Inc, Cupertino CA 95014, USA \\
\email{ashutosh@brainoft.com}
}
\date{Received: date / Accepted: date}

\maketitle

\begin{abstract}
%\boldmath
Human communication takes many forms, including speech, text and instructional videos. It typically has an underlying structure, with a starting point, ending, and certain objective steps between them.  In this paper, we consider instructional videos where there are tens of millions of them on the Internet.  

We propose a method for parsing a video into such semantic steps in an unsupervised way. Our method is capable of providing a semantic ``storyline'' of the video composed of its objective steps. We accomplish this using both visual and language cues in a joint generative model. Our method can also provide a textual description for each of the identified semantic steps and video segments. We evaluate our method on a large number of complex YouTube videos and show that our method discovers
semantically correct instructions for a variety of tasks.
\footnote{First version of this paper appeared in ICCV 2015. This extended version has more details
on the learning algorithm and hierarchical clustering with full derivation, additional analysis on the robustness to the subtitle noise, and a novel application on robotics.}
\end{abstract}

% not want either math or citations to appear in the abstract.


%\todo{
%\begin{itemize}
%\item Over the last decade, there are a large number of unsupervised data such as videos, YouTibe has BLAH... and intelleigne tsystems shuold be making use of this data...... 
%\%item Alyosh Efros journal papers   --re ad first para in intro and related work
%\%item write the paper is orgznied as follows
%\item video --->  atoms  (atom1,atom2) --> actvity cluster
%\item describe data
%\end{itemize}}

% !TEX root = main.tex

\section{Introduction}
In the last decade, we have seen a significant democratization in the way we access and generate information. One of the major shifts has been in moving from expert-curated information sources into crowd-generated large scale knowledge bases such as Wikipedia\cite{wiki}. For example, the way we generate and access \emph{cooking recipes} has been transformed substantially. Google Trends\cite{google_trends} indicates that in the year of 2005 number of Google searches for \emph{cookbooks} were $1.56$ times larger than the number of searches for \emph{cooking videos}. In the year 2016, the number of searches for \emph{cooking videos} is $8.6$ times larger than that of \emph{cookbooks}. This behavior is mostly due to the large volume of cooking videos available on the internet. In an era where an average user gets 2 million videos for the query \emph{How to make a pancake?}, we need computer vision algorithms that can understand such information and represent it to the users in a compact form. Such algorithms are not only useful for humans to digest millions of videos but also useful for robots to learn concepts from online video collections in order to perform tasks.

Considering the intractability of supervised information in large-scale video collections, we believe the key to the unsupervised grounding is utilizing the structural assumptions. Human communication takes many forms, including language and videos. For instance, explaining ``how-to'' perform a certain task can be communicated via language (\eg, Do-It-Yourself books) as well as visual (e.g., instructional YouTube videos) information. Regardless of the form, such human-generated communication is generally structured and has a clear beginning, end, and a set of steps in between. Finding this hidden and objective steps of human communication is a critical step to understand large video collections.

Language and vision provide different, but correlating and complementary information. Challenge lies in that both video frames and language (from subtitles generated via Automatic Speech Recognition) are only a noisy, partial observation of the actions being performed. However, the complementary nature of language and vision gives the opportunity to understand the activities only from these partial observations. In this paper, we present a unified model, considering both of the modalities, in order to parse human activities into activity steps with no form of supervision other than requiring videos to be the same category (\eg, all cooking eggs, changing tires, etc.).

\begin{figure}[h!]
  \includegraphics[width=0.48\textwidth]{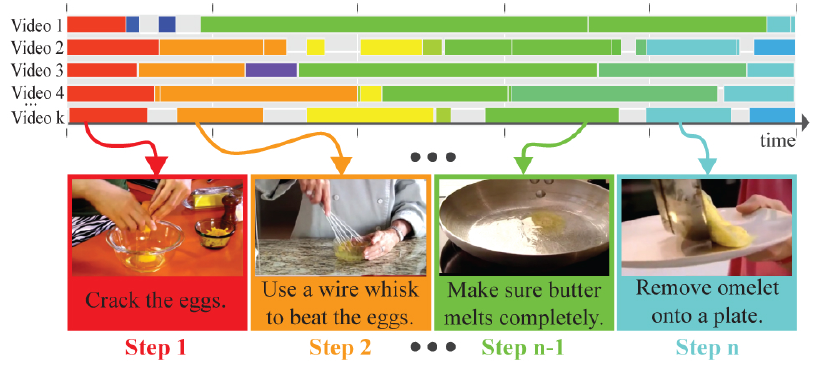}
  \caption{Given a large video collection (frames and subtitles) of an instructional category (\eg, How to cook an ommelette?), we discover activity steps (\eg, crack the eggs). We also parse the videos based on the discovered steps.}
  \label{teaser}
\end{figure}

The key idea in our approach is the observation that the large collection of videos, pertaining to the same activity class, typically include only a few objective activity steps, and the variability is the result of exponentially many ways of generating videos from activity steps through subset selection and time ordering. We study this construction based on the large-scale information available in YouTube in the form of instructional videos  (\eg, ``Making panckage'', ``How to tie a bow tie''). Instructional videos have many desirable properties like the volume of the information and a well defined notion of activity step.  However, the proposed parsing method is applicable to any type of videos as long as they are composed of a set of steps.

The output of our method can be seen as the semantic ``storyline'' of a rather long and complex video collection (see Fig.~\ref{teaser}). This storyline provides what particular steps are taking place in the video collection, when they are occurring, and what their meaning is (\emph{what-when-how}). This method also puts videos performing the same overall task in common ground and capture their high-level relations.

In the proposed approach, given a collection of videos, we first generate a set of language and visual atoms. These atoms are the result of relating object proposals from each frame as well as detecting the frequent words from subtitles. We then employ a generative \emph{beta process mixture model}, which identifies the activity steps shared among the videos of the same category based on a representation using learned atoms. Although we do not explicity enforce this steps to be semantically meaningful, our results highly correlate with the semantic steps. In our method, we do neither use any spatial or temporal label on actions/steps nor any labels on object categories. We later learn a Markov language model to provide a textual description of the activity steps based on the language atoms it frequently uses.

We evaluate our approach on various settings. First of all, we collected a large-scale dataset of instructional videos from YouTube following the most frequently performed \emph{how to} queries. Then, we evaluate temporal parsing quality per video and also a semantic clustering per category (how-to query). Second of all, we extensively analyze the contribution of each modality as well as the robustness against the language noise. Robustness against the language noise is a critical one since ASR always expected to have some errors. Moreover, results suggest that both language and vision is critical for semantic parsing. Finally, we discuss and present a novel robotics application. We start with a single query and generate a detailed physical plan to perform the task. We present a compelling simulation results suggesting that our algorithm has a great potential for robotics applications. 

\section{Related Work}
\label{related}
Designing an artificial intelligence agent which can understand human generated videos have been topic of computer vision and robotics researchers for decades. Motivated by the application of surveillance, video summarization was one of the earliest methods which are related to our problem. The surveillance applications further motivated the activity and event recognition methods. With the help of the availability of larger datasets, researchers managed to train machine learning models which can detect certain events. Recently, the datasets have gotten larger and cross-modal enabling algorithms which can link vision with language. In the mean time, the focus of robotics community was on parsing recipes directly for manipulation. We list and discuss related works from each field in the following sections.

\subsection{Video Summarization:}
Summarizing an input video as a sequence of key frames (static) or video clips (dynamic) is useful for both multimedia search interfaces and retrieval purposes. Early works in the area are summarized in~\cite{vidAbstraction} and mostly focus on \emph{choosing keyframes}.
%Keyframes are also improved by using the video tags and spatio-temporal information~\cite{beyondSearch,createSum}.
%% AMIR: put a '~' instead of ' ' before \cite

%%AMIR: could be briefed
Summarizing videos is particularly important for some specific domains like ego-centric videos and news reports as they are generally long in duration. There are many succesful works~\cite{lee2012discovering, lu2013story,rui2000automatically}; however, they mostly rely on characteristics specific to the domain.

%http://www.sangminoh.org/Publications_files/Oh2012bmvc_videography.pdf --> learn atomic representations and do not do a segmentation
Summarization is also applied to the large image collections by recovering the temporal ordering and visual similarity of images \cite{storyGraph}, and by Gupta et al. ~\cite{gupta2009understanding} to videos in a supervised framework using action annotations. These collections are also used for key-frame selection \cite{khosla2013large} and further extended to video clip selection \cite{kim2014joint,potapov2014category}. Unlike all of these methods which  focus on forming a set of key frames/clips for a compact summary (which is not necessarily semantically meaningful), we provide a fresh approach to video summarization by performing it through semantic parsing on vision and language. However, regardless of this dissimilarity, we experimentally compare our method against them.

\subsection{Modeling Visual and Language Information:}
Learning the relationship between the visual and language data is a crucial problem due to its immense applications. Early methods \cite{matching} in this area focus on learning a common multi-modal space in order to jointly represent language and vision. They are further extended to learning higher level relations between object segments and words \cite{connecting}. Similarly, Zitnick et al.\cite{zitnick2013learning,zitnick2013bringing} used abstracted clip-arts to understand spatial relations of objects and their language correspondences. Kong et al. \cite{kong2014you} and Fidler et al. \cite{fidler2013sentence} both accomplished the task of learning spatial reasoning by only using the image captions. Relations extracted from image-caption pairs, are further used to help semantic parsing \cite{yu2013grounded} and activity recognition \cite{motwani2012improving}. Recent works also focus on automatic generation of image captions with underlying ideas ranging from finding similar images and transferring their captions \cite{ordonez2011im2text} to learning language models conditioned on the image features \cite{kiros2014multimodal,socher2014grounded,farhadi2010every}; their employed approach to learning language models is typically either based on graphical models \cite{farhadi2010every} or neural networks \cite{socher2014grounded,kiros2014multimodal,deepAlignment}.

All aforementioned methods are using supervised labels either as strong image-word pairs or weak image-caption pairs, while our method is fully unsupervised.

\subsection{Activity/Event Recognition:}
The literature of activity recognition is broad. The closest techniques to ours are either supervised or focus on detecting a particular (and often short) action in a weakly/unsupervised manner. Also, a large body of action recognition methods are intended for trimmed videos clips or remain limited to detecting very short actions~\cite{kuehne2011hmdb, UCF101, niebles10_eccv, laptev08_cvpr, efros03_iccv, ryoo09_iccv}. Even though some recent works attempted action recognition in untrimmed videos~\cite{THUMOS14, oneata2014lear, jainuniversity}, they are mostly fully supervised.

Additionally, several method for localizing instances of actions in rather longer video sequences have been developed~\cite{duchenne09_iccv, hoai11_cvpr, laptev07_iccv, bojanowski14_eccv, pirsiavash14_cvpr}. Our work is different from those in terms of being multimodal, unsupervised, applicable to a video collection, and not limited to identifying predefined actions or the ones with short temporal spans.
%Representing actions as 2D+t tubes is a common strategy for action recognition~\cite{blank05_iccv, brendel11_iccv, ma13_iccv}. Recently, there are works that use hierarchical spatiotemporal segments to capture the multi-scale characteristics of actions \cite{brendel11_iccv, ma13_iccv}. Our representation differs in that we can discover the action-related spatiotemporal segments by joint processing the video collection from a pool of proposals.
Also, the previous works on finding action primitives such as~\cite{niebles10_eccv, yao10b_cvpr, jain13_cvpr,lan14_eccv, lan14_vs} are primarily limited to discovering atomic sub-actions, and therefore, fail to identify complex and high-level parts of a long video.

Recently, event recounting has attracted much interest and intends to identify the evidential segments for which a video belongs to a certain class~\cite{sun2014discover,das2013thousand,barbu2012video}. Event recounting is a relatively new topic and the existing methods mostly employ a supervised approach. Also, their end goal is to identify what parts of a video are highly related to an event, and not parsing the video into semantic steps.
%This setup is similar to Hughes et al.\cite{npActivity}. However, we differ in the choices of the underlying distributions since we based our model on semantic multi-modal information.

% AMIR: flow graph sounds like 'storyline'. Consider replacing/removing it.
%Mori et al.\cite{flowGraph} also learns the relations of the actions in terms of a flow graph with the help of a supervision.

\subsection{Recipe Understanding:}
Following the interest in community generated recipes in the web, there have been many attempts to automatically process recipes. Recent methods on natural language processing \cite{cookingSemantics,logicRecipe} focus on semantic parsing of language recipes in order to extract actions and the objects in the form of predicates. Tenorth et al.\cite{logicRecipe} further process the predicates in order to form a complete logic plan. The aforementioned approaches focus only on the language modality and they are not applicable to the videos. The recent advances \cite{beetz,cookie} in robotics use the parsed recipe in order to perform cooking tasks. They use supervised object detectors and report a successful autonomous experiment. In addition to the language based approaches, Malmaud et al.\cite{alignment} consider both language and vision modalities and propose a method to align an input video to a recipe. However, it can not extract the steps automatically and requires a ground truth recipe to align. On the contrary, our method uses both visual and language modalities and extracts the actions while autonomously discovering the steps. There is also an approach which generates multi-modal recipes from expert demonstrations \cite{photoshop}. However, it is developed only for the domain of "teaching user interfaces" and are not applicable to videos.

In summary, three aspects differentiate this work from the majority of existing techniques: 1) discovering semantic steps from a video category, 2) being unsupervised, 3) adopting a multi-modal joint vision-language model for video parsing.
% !TEX root = main.tex

\section{Problem Overview}
\label{sec:overview}
\begin{figure*}[h]
  \includegraphics[width=\textwidth]{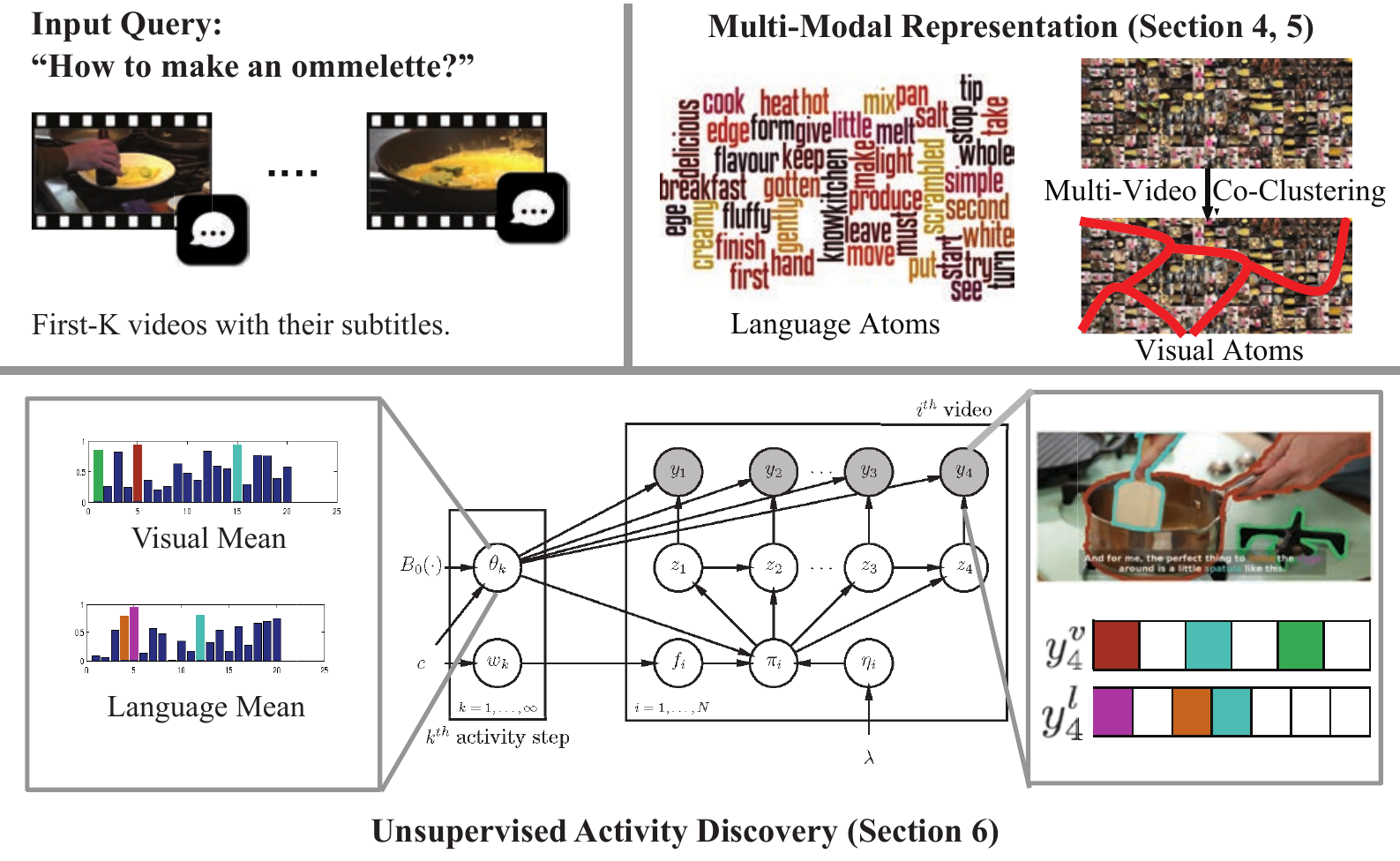}
  \caption{\textbf{Summary of our method.} We start with a single natural language query like \emph{how to make an omelette} and then we crawl the top $K$ videos returned by this query from YouTube. We learn a multi-modal dictionary composed of salient words and object proposals. Rest of the algorithm represents frames and activities in terms of the learned dictionary. For example, in the bottom figure, colors represent such atoms and both activity descriptions $\Theta$ and frame representations $y_t$ are defined in terms of these atoms. (see Fig~\ref{teaser} for output)}
\label{fig:overview}
\end{figure*}

Our algorithm takes an \emph{how-to} sentence as an input query which we further use to download a large-collection of videos. We then learn a multi-modal dictionary using a novel hierarchical clustering approach. We finally use the learned dictionary in order to discover and localize activity steps. We visualize this process in Figure~\ref{fig:overview} with a toy example. The output of our algorithm is temporal parsing of each video as well as an id for each semantic activity step. In other words, we not only temporally segment each video, we also relate the occurrence of same activity over multiple videos with each other. We further visualize the output in Figure~\ref{teaser}.

\smallskip
\noindent\emph{Atoms:} Given a large video-collection composed of visual information as well as subtitles, our algorithm starts with learning a set of visual and language \emph{atoms} which are further used for representing multimodal information (Section~\ref{atoms}). These atoms are designed to be likely to correspond to the mid-level semantic concepts like actions and objects. In order to learn language \emph{atoms}, we find frequently occurring salient words among the subtitles using tf-idf like approach. Learning visual atoms is slightly trickier due to the intra-cluster variability of visual concepts. We generate object proposals and jointly-cluster them into mid-level atoms to obtain visual atoms. We develop a hierarchical clustering algorithm for this purpose (Section~\ref{jointProp}).

\smallskip
\noindent\emph{Discovering Activities:} After learning the atoms, we represent the multi-modal information in each frame based on the occurrence statistics of the atoms. Given the multi-modal representation of each frame, we discover set of temporal clusters occurring over multiple videos using a non-parametric Bayesian method (Section~\ref{learning}). We expect these clusters to correspond to the activity steps which construct the high level activities. Our empirical results confirms this as the resulting clusters significantly correlates with the semantic activity steps.

%In , we explain how we learn the language and visual atoms. Then, we explain the multi-modal representation of the frames in . Finally we explain how do we discover activity steps in 

%In this section, we explain the high-level components of our method which we visualize in Figure~\ref{fig:overview}. Our proposed method consists of three major components; \textbf{(1) Online query and filtering:} Our system starts with querying the YouTube with an \emph{How to} question, and records the top 100 resulting videos. In order to detect the similarity of the videos quickly, we also process the text descriptions and eliminate outliers. \textbf{(2) Frame-wise multi-modal representation:} In order to semantically represent the spatio-temporal information in the videos, we process both the visual and language content of each video. We extract the region proposals and jointly cluster them to detect semantic visual objects. We also detect the salient words of the subtitles. Finally, we epresent the each frame in terms of the resulting objects and the salient words. \textbf{(3) Unsupervised joint clustering:} After describing the each frame by using both language and visual cues, we apply 

% !TEX root = main.tex

\setlength{\tabcolsep}{1mm}
\begin{table}
\footnotesize
\centering
\caption{Notation of the Paper}
\resizebox{1\columnwidth}{!}{%
\begin{tabular}{r|c}
\multicolumn{2}{c}{Learning Atoms} \\
\toprule
$I_t$ & $t^{th}$ frame of the video \\
 $L_t$ & subtitle for $t^{th}$ frame \\
$y_t$ = $\left[y_t^v,y_t^l\right]$ & feature representation of $t^{th}$ frame \\
 $x^p_{i,r}$ & $1$ if $p^{th}$ cluster has $r^{th}$ proposal of $i^{th}$ video, $0$ o.w.  \\
 $z_t$ & activity ID of frame  $t$  \\
\bottomrule
\multicolumn{2}{c}{} \\
\multicolumn{2}{c}{Learning Activities - Beta Process HMM} \\
\toprule 
$x^p$ & binary vector for $p^{th}$ cluster\\
$f_i^k$ & $1$ if $i^{th}$ video has $k^{th}$ activity $0 o.w.$ \\
$\Theta_k$ = $\left[\Theta_k^v,\Theta_k^l\right]$ & emission prob. of $k^{th}$ activity \\
$\eta_i^{k,k^\prime}$ & $P(z_{t+1}=k^\prime|z_{t}=k)$ for $i^{th}$ vid \\
$\pi_i^{k,k^\prime}$ & $\eta_i^{k,k^\prime} \times f_i^k \times f_i^{k^\prime}$ \\
\bottomrule
\end{tabular}}
\end{table}
\normalsize

% !TEX root = main.tex

\section{Multi-Modal Representation with Atoms}
\label{atoms}
Finding the set of activity steps over large collection of videos having large visual varieties requires us to represent the semantic information in addition to the low-level visual cues. Hence, we find our language and visual atoms by using mid-level cues like object proposals and frequent words.

\noindent\textbf{Learning Visual Atoms:} In order to learn visual atoms, we create a large collection of object proposals by independently generating object proposals from each frame of each video. These proposals are generated using the Constrained Parametric Min-Cut (CPMC) \cite{cpmc} algorithm based on both appearance and motion cues. We note the $k^{th}$ proposal of $t^{th}$ frame of $i^{th}$ video as $r^{(i),k}_t$. Moreover, we drop the video index $(i)$ if it is clearly implied in the context.

In order to group this object proposals into mid-level visual atoms, we follow a clustering approach. Although any graph clustering approach (eg. Keysegments \cite{keysegments}) can be applied for this, the joint processing of a large video collection requires handling large visual variability among multiple videos. We propose a new method to jointly cluster object proposals over multiple videos in Section~\ref{jointProp}. Each cluster of object proposals correspond to a visual atom.

\noindent\textbf{Learning Language Atoms:}
We define the language atoms as the salient words which occur more often than their ordinary rates based on the \emph{tf-idf} measure. The \emph{document} is defined as the concatenation of all subtitles of all frames of all videos in the collection as $D=\bigcup_{i \in N_C} \bigcup_{t \in T^{(i)}} L_t^i$. Then, we follow the classical tf-idf measure and use it as $tfidf(w,D)=f_{w,D} \times \log \left( 1+ \frac{N}{n_{w}}\right)$ where $w$ is the word we are computing the tf-idf score for, $f_{w,D}$ is the frequency of the word in the \emph{document} $D$, $N$ is the total number of video collections we are processing, and $n_{w}$ is the number of video collections whose subtitle include the word $w$.

We sort words with their "tf-idf" values and choose the top $K$ words as language atoms (\emph{$K=100$ in our experiments}). As an example, we show the language atoms learned for the category \emph{making scrambled egg} in Figure~\ref{fig:overview} %The resulting collection suggests that they correspond to the important objects, actions and adjectives which represent a semantic information occurring over multiple videos.

%\begin{figure}
%\footnotesize
%\emph{sort, place, water, egg, bottom, fresh, pot, crack, cold, cover, time, overcooking, hot, shell, stove, turn, cook, boil, break, pinch, salt, peel, lid, point, high, rules, perfectly, hard, smell, fast, soft, chill, ice, bowl, remove, aside, store, set, temperature, coagulates, yolk, drain, swirl, shake, white, roll, handle, surface, flat}
%\normalsize
%\caption{Language atoms learned for activity class \emph{"How to hard boil an egg?"}}
%\end{figure}

\begin{figure}[h!]
  \includegraphics[width=0.48\textwidth]{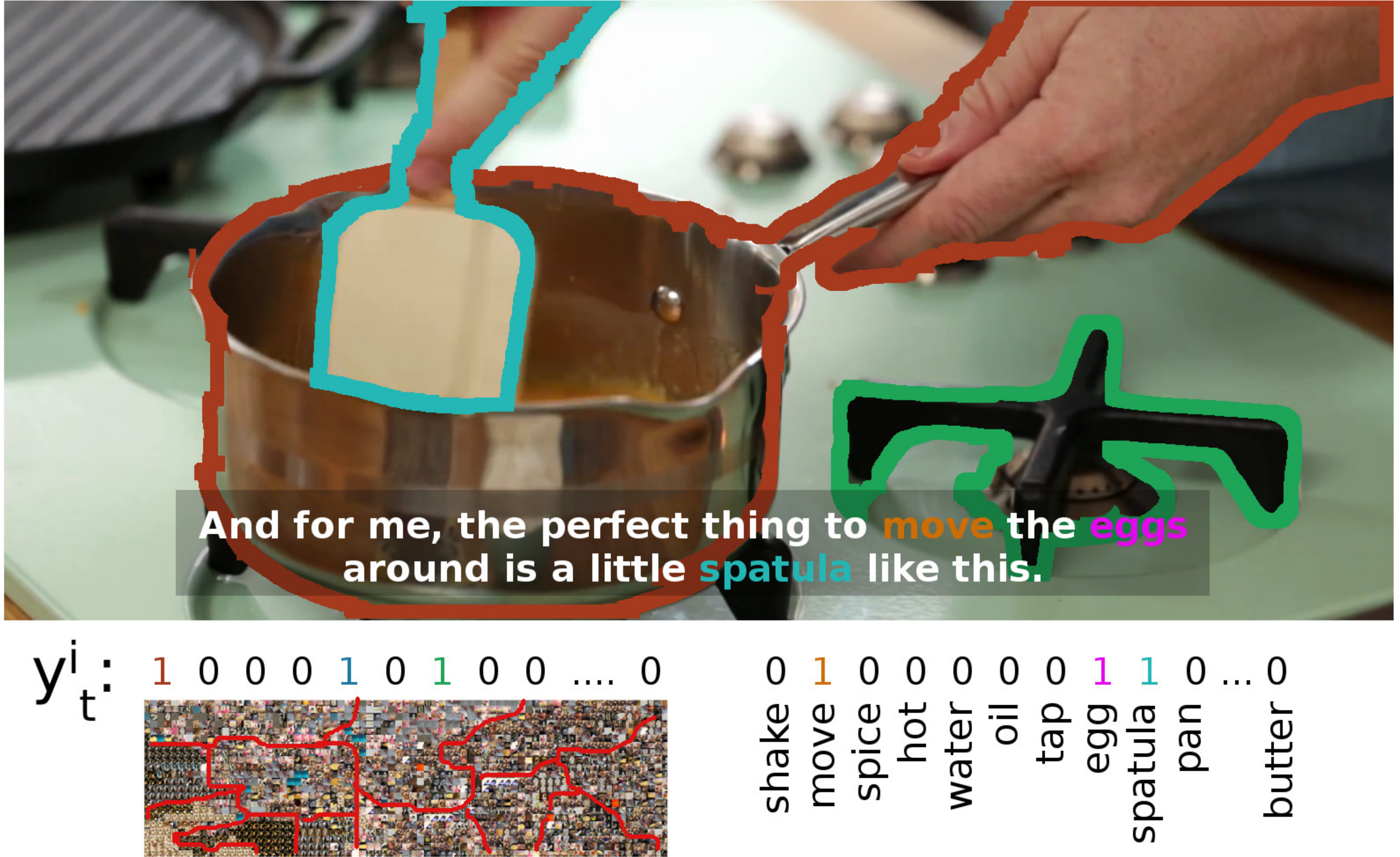}
  \caption{\textbf{Representation for a sample frame.} Three of the object proposals of sample frame are in the visual atoms and three of the words are in the language atoms.}
  \label{visFrame}
\end{figure}

\noindent\textbf{Representing Frames with Atoms:}
After learning the visual and language atoms, we represent each frame via the occurrence of atoms (binary histogram). Formally, the representation of the $t^{th}$ frame of the $i^{th}$ video is denoted as $\mathbf{y^{(i)}_t}$ and computed as \mbox{$\mathbf{y^{(i)}_t}=[\mathbf{y^{(i),l}_t},\mathbf{y^{(i),v}_t}]$} such that $k^{th}$ entry of the $\mathbf{y^{(i),l}_t}$ is $1$ if the subtitle of the frame has the $k^{th}$ language atom and $0$ otherwise. $\mathbf{y^{(i),v}_t}$ is also a binary vector similarly defined over visual atoms. We visualize the representation of a sample frame in the Figure~\ref{visFrame}.

\section{Joint  Clustering over Video Collection}
\label{jointProp}
Given a set of object proposals generated from "multiple videos", simply combining them into a single collection and clustering them into atoms is not desirable for two reasons: (1) semantic concepts have large visual differences among different videos and accurately clustering them into a single atom is hard, (2) atoms should contain object proposals from multiple videos in order to semantically relate the videos. In order to satisfy these requirements, we propose a joint extension to spectral clustering. Note that the purpose of this clustering is generating atoms where each clusters represents an atom.

\begin{figure}[ht]
  \includegraphics[width=0.48\textwidth]{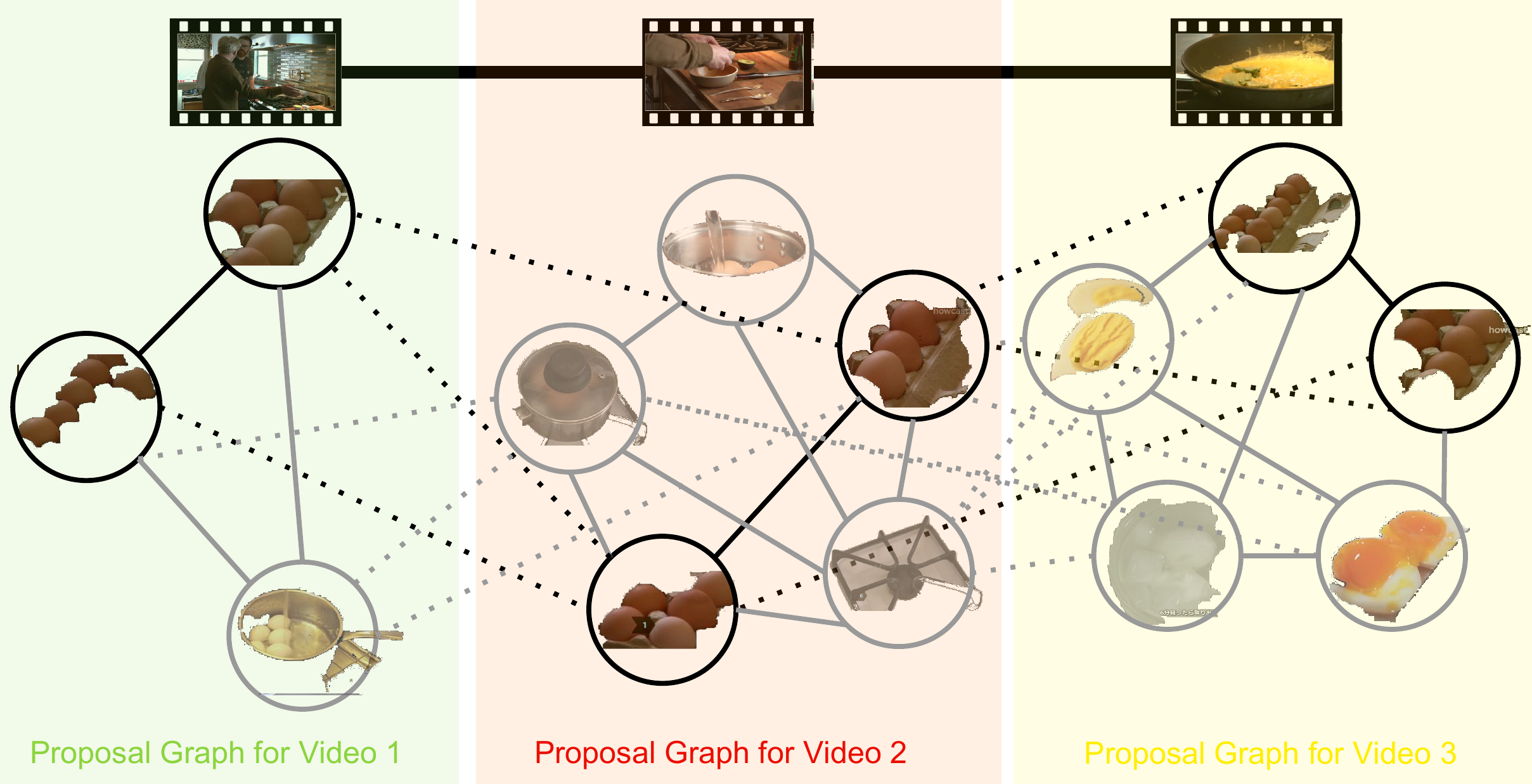}
\caption{\textbf{Joint proposal clustering.} Here, we show the $1^{st}NN$ video graph and $2^{nd}NN$ region graph. Each object proposal is linked to its two NNs from the video it belongs and two NNs from the videos it is neighbour of. Black nodes are the proposals selected as part of the cluster and the gray ones are not selected. Moreover, dashed lines are intra-video edges and solid ones are inter-video edges.}
  \label{hierProposal}
\end{figure}

\noindent\textbf{Basic Graph Clustering:} Consider the set of object proposals extracted from a single video $\{r^k_t\}$, and a pairwise similarity metric $d(\cdot,\cdot)$ for them. We follow the single cluster graph partitioning (SCGP)\cite{scgp} approach to find the dominant cluster which maximizes the intra-cluster similarity:
\begin{equation}
  \argmax_{x^k_t} \frac{\sum_{(k_1,t_1),(k_2,t_2) \in K \times T} x^{k_1}_{t_1} x^{k_2}_{t_2} d(r^{k_1}_{t_1},r^{k_2}_{t_2})}{\sum_{(k,t) \in K \times T} x^{k}_t}
  \label{nonvec}
\end{equation}
where $x^{k}_t$ is a binary variable which is $1$ if $r^{k}_t$ is included in the cluster, $T$ is the number of frames and $K$ is the number of clusters per frame. Adopting the vector form of the indicator variables as $\mathbf{x_{tK+k}}=x^{k}_{t}$ and the pairwise distance matrix as $\mathbf{A}_{t_1K+k_1,t_2K+k_2}=d(r^{k_1}_{t_1},r^{k_2}_{t_2})$, equation (\ref{nonvec}) can be compactly written as
$\argmax_{\mathbf{x}} \frac{\mathbf{x^T}A\mathbf{x}}{\mathbf{x^T}\mathbf{x}}$
This can be solved by finding the dominant eigenvector of $\mathbf{x}$ after relaxing $x^{k}_t$ to $[0,1]$ \cite{scgp,scgp_eigen}. Upon finding the cluster, the members of the selected cluster are removed from the collection and the same algorithm is applied to find remaining clusters.

\smallskip
\noindent\textbf{Joint Clustering:} Our extension of the SCGP into multiple videos is based on the assumption that the key objects occur in most of the videos. Hence, we re-formulate the problem by enforcing the homogeneity of the cluster over all videos.

We first create a kNN graph of the videos based on the distance between their textual descriptions. We use the $\chi^2$ distance of the bag-of-words computed from the video description. We also create the kNN graph of object proposals in each video based on the pretrained "fc7" features of AlexNet\cite{alexnet}. This hierarchical graph structure is visualized in Figure~\ref{hierProposal} for 3 videos sample. After creating this graph, we impose both "inter-video" and "intra-video" similarity among the object proposals of each cluster. Main rationale behind this construction is having a separate notion of distance for inter-video and intra-video relations since the visual similarity decreases drastically for inter-video ones.

Given the intra-video distance matrices $\mathbf{A^{(i)}}$, the binary indicator vectors $\mathbf{x^{(i)}}$, and the inter-video distance matrices as $\mathbf{A^{(i,j)}}$, we define our optimization problem as;
\begin{equation}
\argmax \sum_{i \in N} \frac{\mathbf{x^{(i)^T}}\mathbf{A^{(i)}}\mathbf{x^{(i)}}}{\mathbf{x^{(i)^T}}\mathbf{x^{(i)}}} +
\sum_{i \in N} \sum_{j \in \mathcal{N}(i)} \frac{\mathbf{x^{(i)^T}}\mathbf{A^{(i,j)}}\mathbf{x^{(j)}}} {\mathbf{x^{(i)^T}}\mathds{1}\mathds{1}^T\mathbf{x^{(j)}}},
\end{equation}
where $\mathcal{N}(i)$ is the neighbours of the video $i$ in the kNN graph, $\mathds{1}$ is vector of ones and $N$ is the number of videos.

Although we can not use the efficient eigen-decomposition approach from \cite{scgp,scgp_eigen} as a result of the modification, we can use Stochastic Gradient Descent as the cost function is quasi-convex when relaxed. We use the SGD with the following analytic gradient function:
\begin{equation}
  \nabla_{\mathbf{x^{(i)}}} = \frac{2\mathbf{A^{(i)}} \mathbf{x^{(i)}} -2\mathbf{x^{(i)}} r^{(i)}}
  {\mathbf{{x^{(i)}}^T}\mathbf{x^{(i)}}}
+ \sum_{i \in N} \frac{\mathbf{A^{i,j}}\mathbf{x^{j}} - \mathbf{{x^{(j)}}^T} \mathds{1} r^{(i,j)}}{\mathbf{{x^{(i)}}^T} \mathds{1} \mathds{1}^T \mathbf{x^{(j)}} },
  %\text{Some vector matrix multiplication}
\end{equation}
where $r^{(i)}=\frac{\mathbf{x^{(i)^T}}\mathbf{A^{(i)}}\mathbf{x^{(i)}}}{\mathbf{x^{(i)^T}}\mathbf{x^{(i)}}}$ and $r^{(i,j)}=\frac{\mathbf{x^{(i)^T}}\mathbf{A^{(i,j)}}\mathbf{x^{(j)}}} {\mathbf{x^{(i)^T}}\mathds{1}\mathds{1}^T\mathbf{x^{(j)}}}$

We iteratively use the method to find clusters, and stop after the $K=20$ clusters are found as the remaining object proposals were deemed not relevant to the activity. Each cluster corresponds to a visual atom for our application.

\begin{figure*}[ht]
  \begin{subfigure}[b]{0.32\textwidth}
\includegraphics[width=\textwidth]{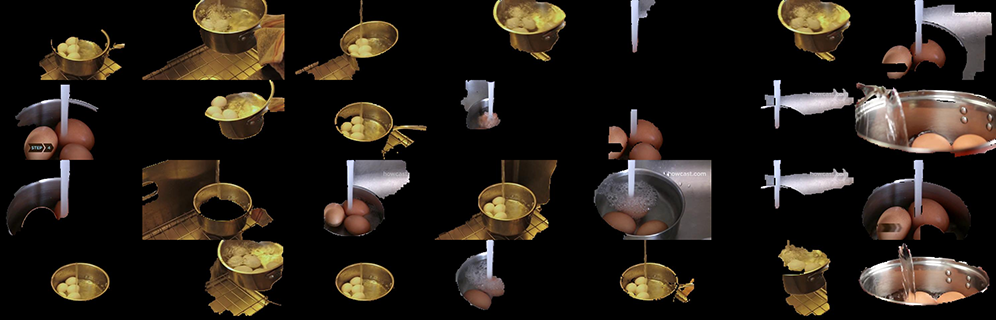}
%\caption{Cluster 1}
\end{subfigure}
~
\begin{subfigure}[b]{0.32\textwidth}
\includegraphics[width=\textwidth]{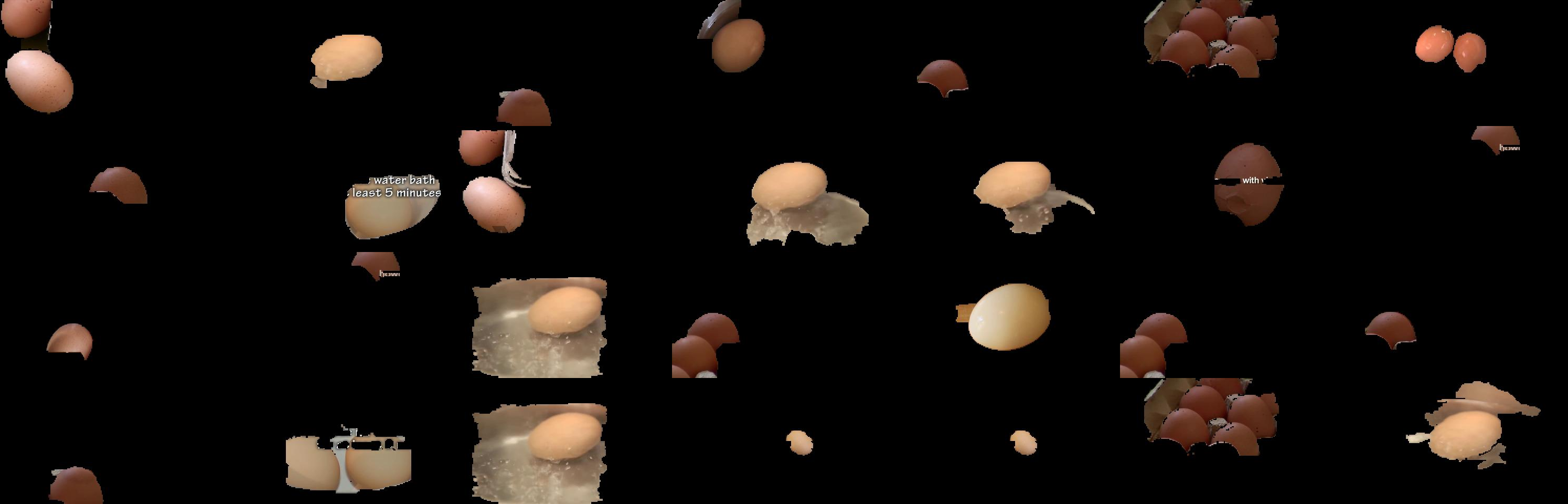}
%\caption{Cluster 2}
\end{subfigure}
~
\begin{subfigure}[b]{0.32\textwidth}
\includegraphics[width=\textwidth]{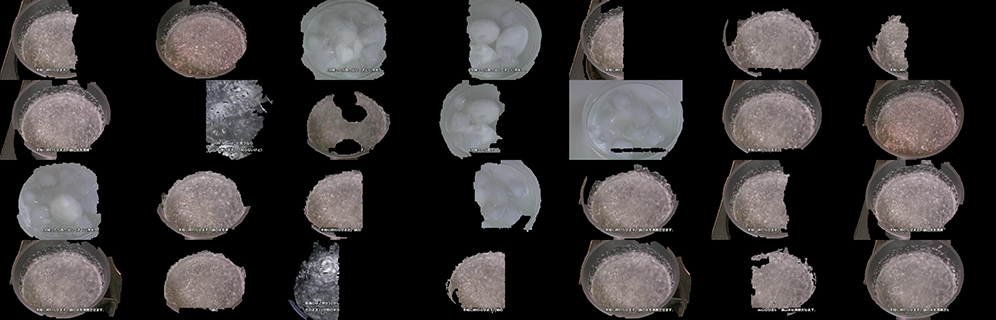}
\end{subfigure}

\begin{subfigure}[b]{0.32\textwidth}
\includegraphics[width=\textwidth]{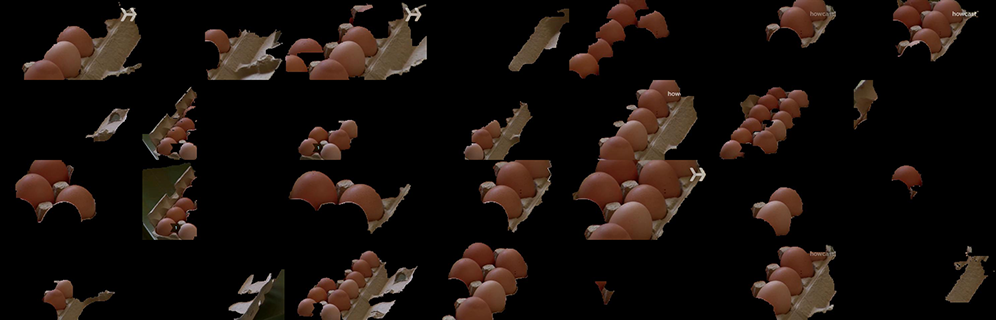}
\end{subfigure}
~
\begin{subfigure}[b]{0.32\textwidth}
\includegraphics[width=\textwidth]{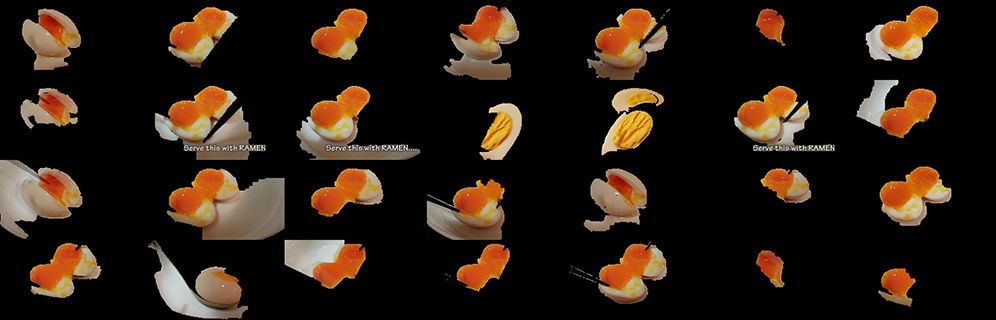}
\end{subfigure}
~
\begin{subfigure}[b]{0.32\textwidth}
\includegraphics[width=\textwidth]{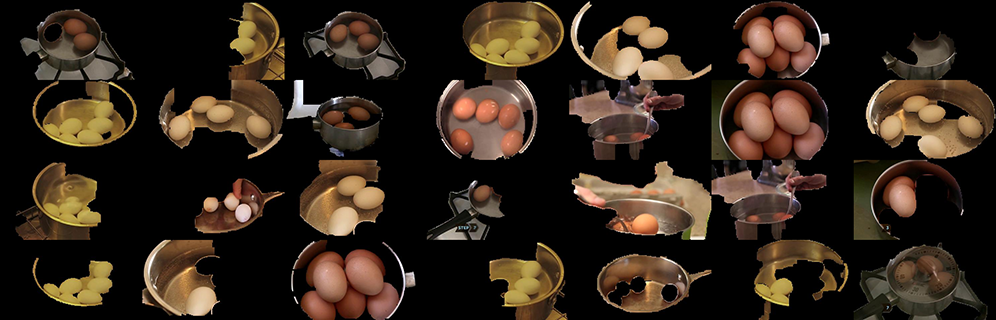}
\end{subfigure}

\begin{subfigure}[b]{0.32\textwidth}
\includegraphics[width=\textwidth]{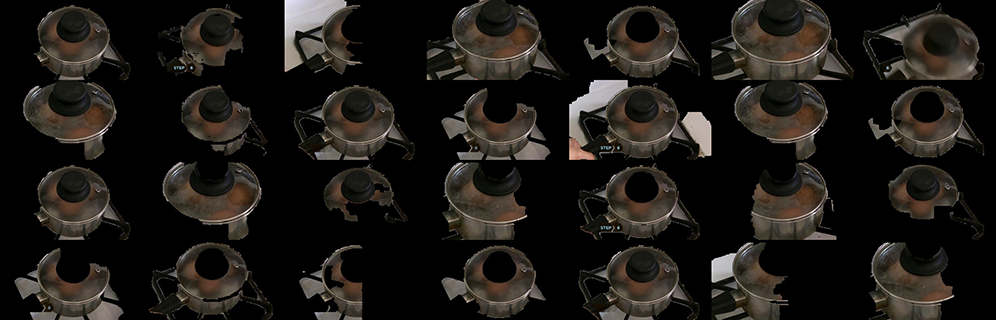}
\end{subfigure}
~
\begin{subfigure}[b]{0.32\textwidth}
\includegraphics[width=\textwidth]{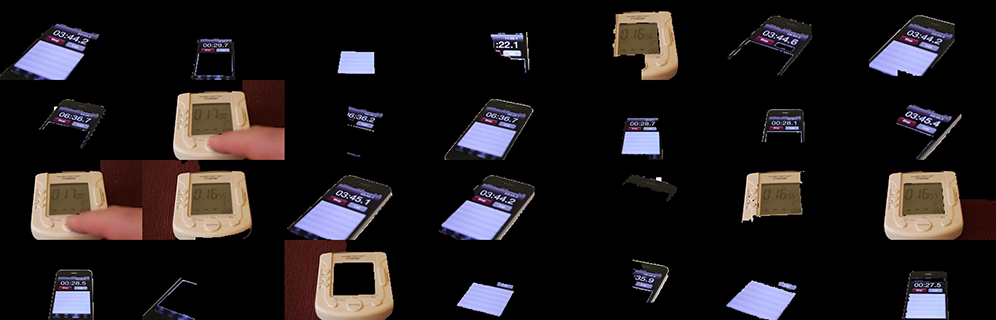}
\end{subfigure}
~
\begin{subfigure}[b]{0.32\textwidth}
\includegraphics[width=\textwidth]{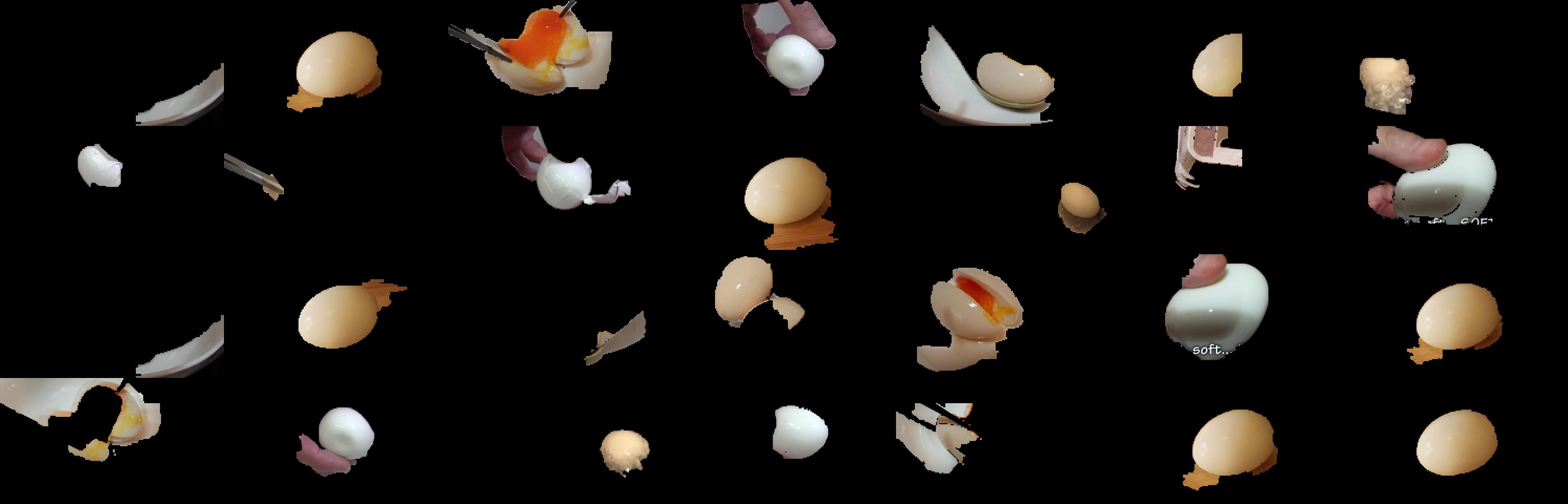}
\end{subfigure}

\caption{\textbf{Randomly selected images of four randomly selected clusters learned for \emph{How to hard boil an egg?}} Please note that the objects in the same cluster is not only coming from a single video but discovered over multiple videos. Hence, this stage helps in linking videos with each other.  Resulting clusters are semantically accurate since they typically belong to a single semantic concept like water filling the pot.}
\label{cvis}
\end{figure*}

In Figure~\ref{cvis}, we visualize some of the atoms (\ie clusters) we learned for the query \emph{How to Hard Boil an Egg?}. As apparent in the figure, the resulting atoms are highly correlated and correspond to semantic objects\&concepts regardless of their significant intra-class variability.

% !TEX root = main.tex

\section{Unsupervised Activity Representation}
\label{basics}
\label{learning}
In this section, we explain our model for discovering the activity steps from a video collection given the language and visual atoms. The main idea behind this step is utilizing the repetitive nature of steps. In other words, although there are large number of videos in any chosen category, the underlying set of steps are very few. Hence, we tried to find smallest set of activities which can generate all the videos we crawl.

We note the extracted representation of the frame $t$ of video $i$ as $\mathbf{y^{(i)}_t}$. We model our algorithm based on activity steps and note the activity label of the $t^{th}$ frame of the $i^{th}$ video as $z^{(i)}_t$. We do not fix the the number of activities and use a non-parametric approach.

%We start with explaining the notation. As we already defined in the previous sections,

In our model, each activity step is represented over the atoms as the likelihood of including them. In other words, each activity step is a Bernoulli distribution over the visual and language atoms as $\theta_k=[\theta_k^l,\theta_k^v]$ such that $m^{th}$ entry of the $\theta_k^l$ is the likelihood of observing $m^{th}$ language atom in the frame of an activity $k$. Similarly, $m^{th}$ entry of the $\theta_k^v$ represents the likelihood of seeing $m^{th}$ visual atom. In other words, each frame's representation $\mathbf{y^{(i)}_t}$ is sampled from the distribution corresponding to its activity as \mbox{$\mathbf{y^{(i)}_t}|z^{(i)}_t=k \sim Ber(\theta_k)$}. As a prior over $\theta$, we use its conjugate distribution -- \emph{Beta distribution} --.

Given the model above, we explain the generative model which links activity steps and frames in Section~\ref{bphmm}.
 %The key idea is each frame is represented via atoms and each activity step is a distribution over atoms hence the atoms are the linkage. We further explain the learning method to fit this model without any supervision in the Section~\ref{gibbs}.

\subsection{Beta Process Hidden Markov Model}
\label{bphmm}
For the understanding of the time-series information, Fox et al.\cite{foxBPHMM} proposed the Beta Process Hidden Markov Models (BP-HMM). In BP-HMM setting, each time-series exhibits a subset of available features. Similarly, in our setup each video exhibits a subset of activity steps.
%It is based on the notion features(\eg activity steps) which can explain the behaviour of a collection of time-series data (\eg video collection).

Our model follows the construction of Fox et al.\cite{foxBPHMM} and differs in the choice of probability distributions since \cite{foxBPHMM} considers Gaussian observations whereas we adopt binary observations of atoms. In our model, each video $i$ chooses a set of activity steps through an activity step vector $\mathbf{f^{(i)}}$ such that $f^{(i)}_k$ is $1$ if $i^th$ video has the activity step $k$, and 0 otherwise. When the activity step vectors of all videos are concatenated, it becomes an activity step matrix $\mathbf{F}$ such that $i^th$ row of the $\mathbf{F}$ is the activity step vector $\mathbf{f^{(i)}}$. Moreover, each activity step $k$ also has a prior probability $b_k$  and a distribution parameter $\theta_k$ which is the Bernoulli distribution as we explained in the Section~\ref{basics}.

In this setting, the activity step parameters $\theta_k$ and $b_k$ follow the \emph{beta process} as;
\begin{equation}
  B|B_0,\gamma,\beta \sim \text{BP}(\beta,\gamma B_o), B=\sum_{k=1}^\infty b_k \delta_{\theta_k}
\end{equation}
where $B_0$ and the $b_k$ are determined by the underlying Poisson process \cite{ibp} and the feature vector is determined as independent Bernoulli draws as $f_{k}^{(i)} \sim Ber(b_k)$. After marginalizing over the $b_k$ and $\theta_k$, this distribution is shown to be equivalent to Indian Buffet Process (IBP)\cite{ibp}. In the IBP analogy, each video is a customer and each activity step is a dish in the buffet. The first customer (video) chooses a $\text{Poisson}(\gamma)$ unique dishes (activity steps). The following customer (video) $i$ chooses previously sampled dish (activity step) $k$ with probability $\frac{m_k}{i}$,  proportional to the number of customers ($m_k$) chosen the dish $k$, and it also chooses $\text{Poisson}(\frac{\gamma}{i})$ new dishes(activity steps). Here, $\gamma$ controls the number of selected activities in each video and $\beta$ promotes the activities getting shared by videos.

The above IBP construction represents the activity step discovery part of our method. In addition, we also need to model the video parsing over discovered steps. Moreover, we need to model this two steps jointly. We model the each video as an Hidden Markov Model (HMM) over the selected activity steps. Each frame has the hidden state --activity step-- ($z^{(i)}_t$) and we observe the multi-modal frame representation $\mathbf{y^{(i)}_t}$. Since we model each activity step as a Bernoulli distribution, the emission probabilities follow the Bernoulli distribution as $p(\mathbf{y^{(i)}_t}|z^{(i)}_t)=Ber(\theta_{z^{(i)}_t})$.

\begin{figure}[h!]
  \includegraphics[width=0.5\textwidth]{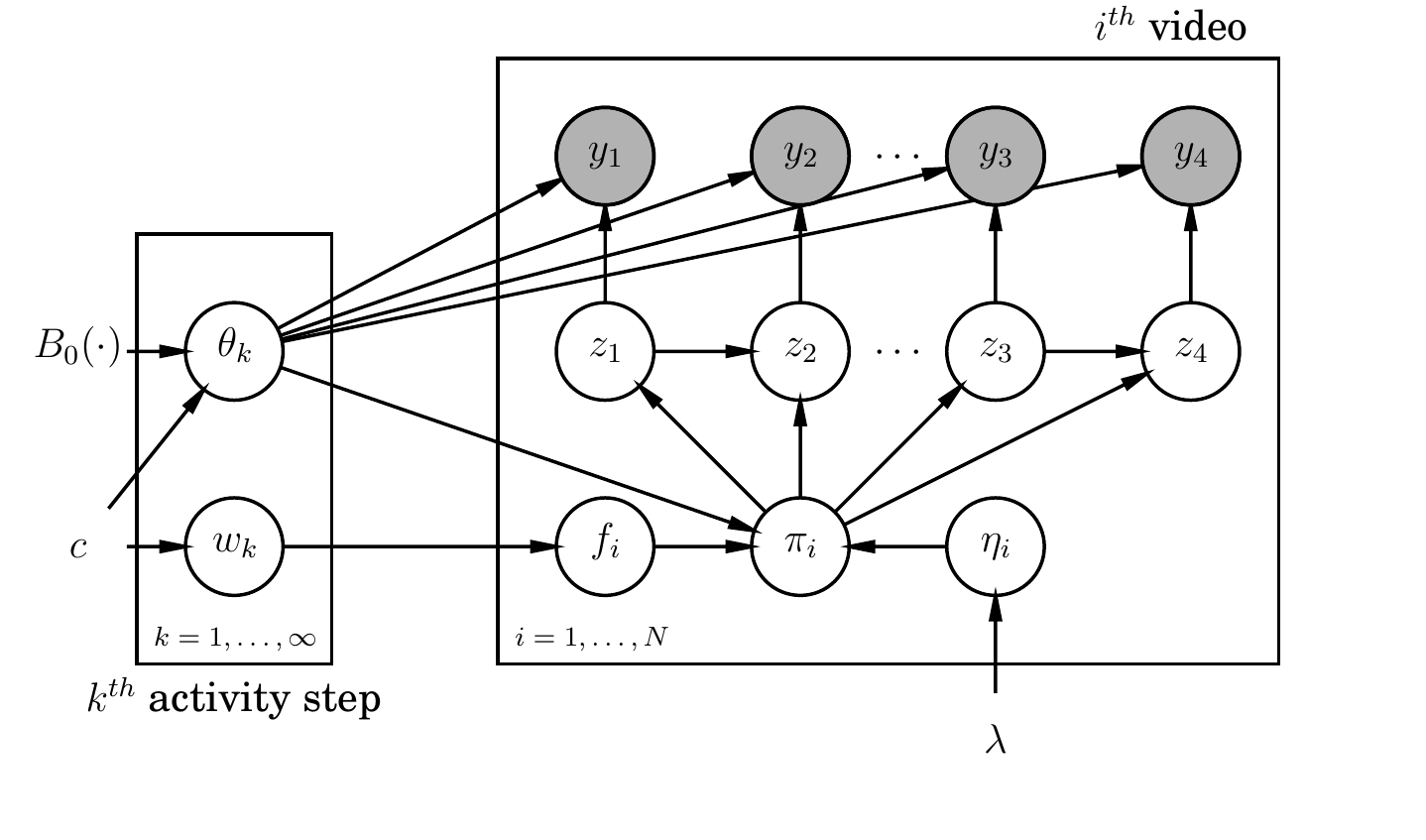}
  \caption{\textbf{Graphical model for BP-HMM:} The left plate represent the activity steps and the right plate represent the videos. \ie the left plate is for the activity step discovery and right plate is for parsing. \emph{See Section~\ref{bphmm} for details.}}
  \label{bphmmo}
\end{figure}

For the transition probabilities of the HMM, we do not put any constraint and simply model it as any point from a probability simplex which can be sampled by drawing a set of Gamma random variables and normalizing them \cite{foxBPHMM}. For each video $i$, a Gamma random variable is sampled for the transition between activity step $j$ and activity step $k$ if both of the activity steps are included by the video (\ie if $f^i_k$ and $f^i_j$ are both $1$). After sampling these random variables, we normalize them to make transition probabilities to sum up 1. This procedure can be represented formally as
\begin{equation}
  \eta_{j,k}^{(i)} \sim Gam(\alpha+\kappa \delta_{j,k},1), \quad \mathbf{\pi_j^{(i)}} = \frac{\mathbf{\eta^{(i)}_j} \circ \mathbf{f^{(i)}}}{\sum_k \eta^{(i)}_{j,k} f^{(i)}_k}
\end{equation}
Where $\kappa$ is the persistence parameter promoting the self state transitions a.k.a. more coherent temporal boundaries, $\circ$ is the element-wise product and $\pi^i_j$ is the transition probabilities in video $i$ from activity step $j$ to other steps. This model is also presented as a graphical model in Figure \ref{bphmmo}

\subsection{Gibbs sampling for BP-HMM}
We employ Markov Chain Monte Carlo (MCMC) method for learning and inference of the BP-HMM. We follow the exact sampler proposed by Fox et al.\cite{foxBPHMM}. It marginalize over activity likelihoods $w$ and activity assignments $\mathbf{z}$ and samples the rest. MCMC procedure iteratively samples the conditional likelihood of activity matrix $\mathbf{F}$, activity parameters $\theta$ and transition weights $\eta$. We divide the explanation of this samplers into two sections, sampling the activities through activity matrix ($\mathbf{F}$) and activity parameters ($\theta$), and sampling the HMM parameters $\mathbf{\eta}$. Marginalization over activity assignments follows the efficient dynamic programming approach.

\paragraph{Sampling the Activities:}
Consider the binary activity inclusion matrix $\mathbf{F}$ such that $F_{i,j}$ is $1$ if the $i^{th}$ video has the $j^{th}$ activity. Following the sampler of Fox et al.\cite{foxBPHMM}, we divide the sampling $\mathbf{F}$ into two parts, namely, sampling the shared activities and sampling the novel activities. Sampling shared activities correspond the re-sampling of existing entries of $\mathbf{F}$. We simply iterate over each entry and propose a flip (\ie if the $i^{th}$ video has the $j^{th}$ activity, we propose to flip it and not to include $j^{th}$ activity in the $i^{th}$ video). We accept or reject this proposals following the Metropolis-Hasting rule.

In order to sample the novel activities, we follow the data-driven sampler \cite{npActivity}. Consider the case in which we want to propose a novel activity by setting the $F_{i,j+1}$ to $0$. In other words, we introduce a new activity ($j+1^{th}$ activity) such that $i^{th}$ video includes it. In order to sample the parameters $\theta_{j+1}$ of it, we first sample a temporal window $W$ over the $i^{th}$ video. This window is sampled by sampling the starting frame and the length of the window from a uniform distribution. Then, we sample the novel activity from \emph{Beta distribution} as;
\begin{equation}
  \theta_{k,n}|W \sim \text{Beta}(\alpha_n,\beta_n)
\end{equation}
where $\theta_{k,n}$ is the $n^{th}$ entry of $\theta_k$, $\alpha_n$ is the number of frames in the window $W$ which have the atom $n$, and $\beta_n$ is the number of frames which do not have the atom $n$. We use \emph{Beta distribution} because it is the conjugate prior of the \emph{Bernoulli distribution} that we use to model activities.

\paragraph{Sampling the HMM Parameters:}
When the activities are defined via $\Theta$ and each video selects a subset of them via ($\mathbf{F}$), we can compute the likelihood of each state assignment by using the dynamic programming given the transition probabilities $\mathbf{\eta}$. By using the likelihoods, we sample the state assignments $\mathbf{z}$.

When the states are sampled, we can use the closed-form sampler derived in \cite{foxBPHMM}. Fox et al.\cite{foxBPHMM} shows that the transition probabilities can be sampled through a \emph{Dirichlet} random variable and scaling it with a \emph{Gamma} random variable as;
\begin{equation}
  \mathbf{\pi^{(i)}} \sim Dir(\ldots,N^{(i)}_{j,k}+\alpha+\delta_{j,k}\kappa,\ldots)
\end{equation}

followed by \mbox{$\mathbf{\eta^{(i)}}=\mathbf{\pi^{(i)}} \times C^{(i)}$} such that 

\noindent \mbox{$C^{(i)} \sim Gamma( K^{(i)}_+\lambda +\kappa,1)$}. Here, $N^{(i)}_{j,k}$ represents the number of transitions between state $j$ and state $k$ in the video $i$, $\alpha$, $\lambda$ and $\kappa$ are hyperparameters which we learn with cross-validation, $\delta_{j,k}$ is $1$ if $j=k$ and $0$ o.w., and $K^{(i)}_+$ is the number of activities the $i^{th}$ video has chosen.

At the end of the Gibbs sampling, our algorithm ends with a set of activities each represented with respect to the discovered atoms \ie $ \Theta_1 \ldots \Theta_k$ and label of each frame among the discovered activities $[1,\ldots, k]$. $\Theta_i$ can be considered as a generative distribution of each discovered activity. In other words, if we want to sample a frame from activity $i$, we simply sample set of language and visual atoms from $\Theta_i$. We perform this sampling in order to generate a language caption for each discovered activity as explained in Section~\ref{langgen}. We also consistently visualize the results of discovery using story lines as shown in Figure~\ref{teaser}. We assign a color code to each discovered activity and sample keyframes from 4 four different clips of same activity. We further generate a natural language description as well as display the temporal segmentation of each video as a colored timeline.
% !TEX root = main.tex

\section{Experiments}
In order to experiment the proposed method, we first collected a dataset (details in Section~\ref{dataset:sec}). We labelled small part of the dataset with frame-wise activity step labels and used it as an evaluation corpus. Neither the set of labels, nor the temporal boundaries are exposed to our algorithm since the set-up is completely unsupervised. We evaluate our algorithm against the several unsupervised clustering baselines and state-of-the-art algorithms from video summarization literature which are applicable.

\subsection{Dataset}
\label{dataset:sec}
We use WikiHow\cite{wikiHow} in order to obtain the top100 queries the internet users are interested in and choose the ones which are
directly related to the physical world. Resulting queries are;

\emph{\textbf{How to}}\footnotesize
\emph{Bake Boneless Skinless Chicken, Make Jello Shots, Cook Steak, Bake Chicken Breast, Hard Boil an Egg, Make Yogurt, Make a Milkshake, Make Beef Jerky, Tie a Tie, Clean a Coffee Maker, Make Scrambled Eggs, Broil Steak, Cook an Omelet, Make Ice Cream, Make Pancakes, Remove Gum from Clothes, Unclog a Bathtub Drain}
\normalsize

For each of the queries, we crawled YouTube and got the top 100 videos. We also downloaded the English subtitles if they exist. We further randomly choose 5 videos out of 100 per query. Although the choice was random, we discarded outlier videos at this stage and re-sampled without replacement to have 5 inlier evaluation video per query. Other than outlier removal, no human super vision is used to choose evaluation videos. Hence, we have total of 125 evaluation videos and 2375 unlabelled videos.

For each evaluation video, we asked an independent labeler to label them. The dataset is labeled by 5 independent labelers each annotating 5 categories. We asked labelers to label start and end frame of each activity step as well as the name of the step. We simply asked them the question \emph{What are the activity steps and where does each of starts and end?}. All labelers are shown 5 wikiHow\cite{wikihow} video recipes with detailed steps before starting the annotation process as a baseline. 

\subsubsection{Outlier Video Removal}
\begin{figure}[ht]
%  \begin{subfigure}[b]{0.5\textwidth}
    \includegraphics[width=0.5\textwidth]{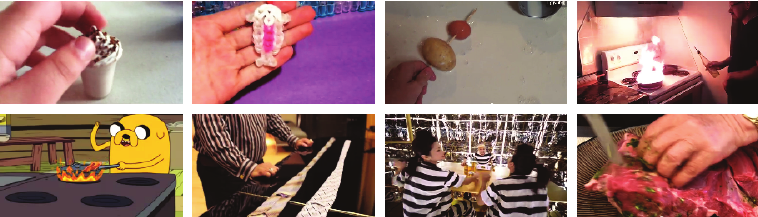}
%  \end{subfigure}~
\caption{\textbf{Sample videos which our algorithm discards as an outlier for various queries.}
A toy milkshake, a milkshake charm, a funny video about How to NOT make smoothie, a video about the danger of a fire, a cartoon video, a neck-tie video erroneously labeled as bow-tie, a song, and a lamb cooking mislabeled as chicken.}
\label{outliers}
\end{figure}

% \label{filter} 
The video collection we obtain without any expert intervention might have outliers; since, our queries are typical daily activities and there are many cartoons, funny videos, and music videos about them. Hence, we have an automatic filtering stage. The key-idea behind the filtering algorithm is the fact that instructional videos have a distinguishable text descriptions when compared with outliers. To exploit this, we use a clustering algorithm to find the large cluster of instructional videos with no outlier. Given a large video collection, we use the graph we explain in Section~\ref{jointProp} and compute the dominant video cluster by using the Single Cluster Graph Partitioning \cite{scgp} and discards the remaining videos as outlier. We represent each video as a bag-of-words of their textual description. In Figure~\ref{outliers}, we visualize some of the discarded videos. Although our algorithm have false positives while detecting outliers, we always have enough number of videos (minimum 50) after the outlier detection thanks to the large-scale dataset.

\subsection{Qualitative Results}
\begin{figure*}[ht]
  \begin{subfigure}[b]{\textwidth}
    \includegraphics[width=\textwidth]{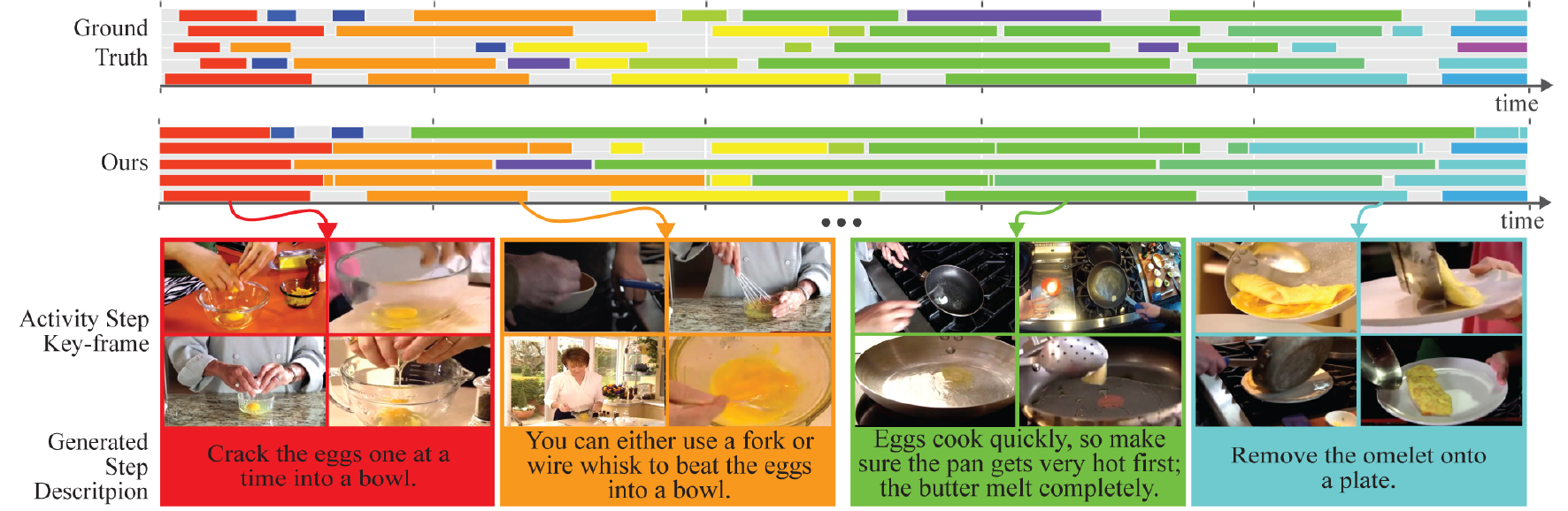}
    \caption{How to make an omelet?}
    \label{recipe:ommelette}
  \end{subfigure}

  \begin{subfigure}[b]{\textwidth}
    \includegraphics[width=\textwidth]{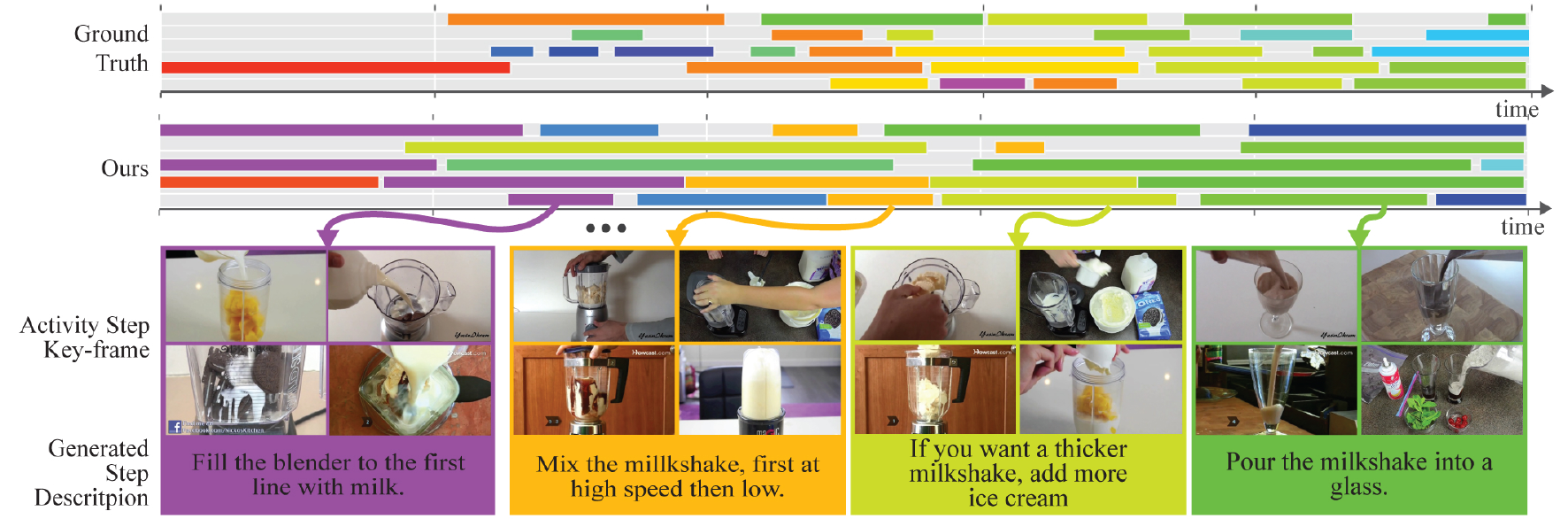}
    \caption{How to make a milkshake?}
    \label{recipe:milkshake}
  \end{subfigure}~
\caption{\textbf{Video storylines for queries \emph{How to make an omelet?} and \emph{How to make a milkshake?}} Temporal segmentation of the videos and ground truth segmentation. We also color code the activity steps we discovered and visualize their key-frames and the automatically generated captions. \emph{Best viewed in color.}}
\label{recipe:overall}
%\normalsize}
\end{figure*}

After independently running our algorithm on all categories, we discover activity steps and parse the videos according to discovered steps. We visualize some of these categories qualitatively in Figure~\ref{recipe:overall} with the temporal parsing of evaluation videos as well as the ground truth parsing.

To visualize the content of each activity step, we display key-frames from different videos. We also train a 3rd order Markov language model\cite{languageModel} by using the subtitles. Moreover, we generate a caption for each activity step by sampling this model conditioned on the $\theta^l_k$. We explain the details of this process in the appendix.

As shown in the Figures~\ref{recipe:ommelette}\&\ref{recipe:milkshake}, resulting steps are semantically meaningful. Moreover, the language captions are also quite informative hence we can conclude that there is enough language context within the subtitles in order to detect activities. On the other hand, some of the activity steps always occur together and our algorithm merges them into a single step while promoting sparsity.

\subsection{Quantitative Results}
We compare our algorithm with the following baselines.

\vspace{3mm}
\noindent\textbf{Low-level features (LLF):}
In order to experiment the effect of learned atoms, we compare with low-level features. As features, we use the state-of-the-art Fisher vector representation of HOG, HOF and MBH features \cite{THUMOS14}.

\vspace{3mm}
\noindent\textbf{Single modality:}
To experiment the effect of multi-modal approach, we compare with single modality approach by only using the atoms of a single modality.

\vspace{3mm}
\noindent\textbf{Hidden Markov Model (HMM):}
To experiment the effect of joint generaive model, we compare our algorithm with an HMM. We use the Baum-Welch \cite{rabiner} with cross-validation.

\vspace{3mm}
\noindent\textbf{Kernel Temporal Segmentation\cite{potapov2014category}:}
Kernel Temporal Segmentation (KTS) proposed by Potapov et al.\cite{potapov2014category} can detect the temporal boundaries of the events/activities in the video from a time series data without any supervision. It enforces a local similarity of each resultant segment.

\begin{figure*}[t]
  \includegraphics[width=\textwidth]{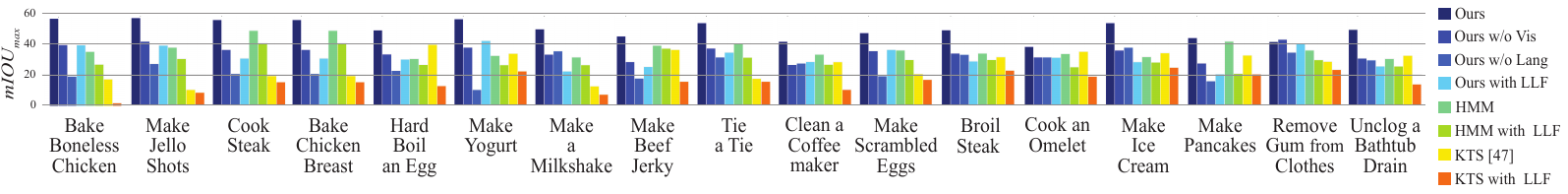}
  \caption{\textbf{$IOU_{max}$ values for all categories, for all competing algorithms.} Results suggest that our algorithm is outperforming all other baselines. It also suggests that the visual information is contributing more than language for temporal intersection over union. This is rather expected since people tend to talk about things they did and will do; hence, language is expected to have low localization accuracy.}
  \label{mIOU}
\end{figure*}
\begin{figure*}
\includegraphics[width=\textwidth]{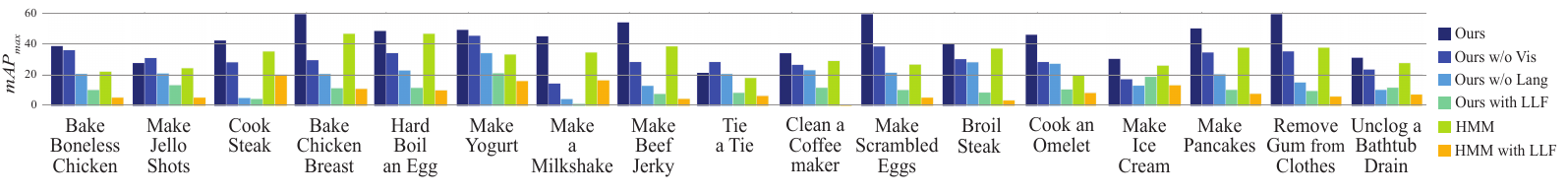}
\caption{\textbf{$AP_{max}$ values for all categories, for all competing algorithms.} Results suggest that our algorithm is outperforming all other baselines in most of the cases. The failure cases included recipes like \emph{How to tie a tie?} which is rather expected since a video about tying a tie only includes a tie in the scene which is not informative enough to distinguish steps. The results also suggest that language contributes more than visual information for average precision, which is also rather expected since same step has very high visual variance and generally referred by using same or similar words.}
\label{mmAP}
\end{figure*}

Given parsing results and the ground truth, we evaluate both the quality of temporal segmentation and the activity step discovery. We base our evaluation on two widely used metrics; intersection over union ($IOU$) and mean average precision($mAP$). \noindent$\mathbf{IOU}$ measures the quality of temporal segmentation and it is defined as; $\frac{1}{N}\sum_{i=1}^N \frac{\tau^\star_i \cap \tau^\prime_{i}}{\tau^\star_i \cup \tau^\prime_{i}}$ where $N$ is the number of segments, $\tau^\star_i$ is ground truth  segment and $\tau^\prime_{i}$ is the detected segment. \noindent$\mathbf{mAP}$ is defined per activity step and can be computed based on a precision-recall curve \cite{THUMOS14}. In order to adopt these metrics into unsupervised setting, we use cluster similarity measure(csm)\cite{liao05} which enables us to use any metric in unsupervised setting. It chooses a matching of ground truth labels with predicted labels by searching over all matching and choosing the ones giving highest score. We use $mAP_{csm}$ and $IOU_{csm}$ as evaluation metrics.

\begin{figure*}[t]
  \includegraphics[width=\textwidth]{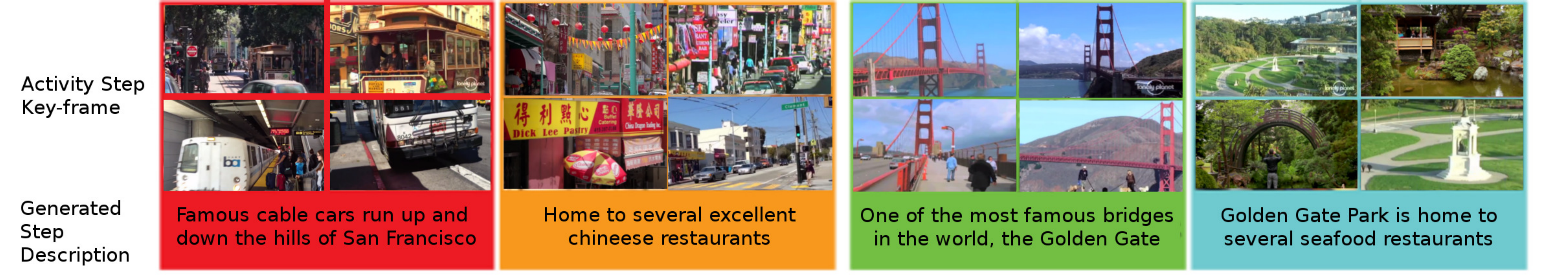}
  \caption{\textbf{Qualitative results for parsing `Travel San Francisco' category.} The results suggest that our algorithm can generalize categories beyond instructional videos. For example, travel videos can also be parsed using our method.}
  \label{sf}
\end{figure*}

\paragraph{\textbf{Accuracy of the temporal parsing.}}
We compute, and plot in Figure\ref{mIOU}, the $IOU_{cms}$ values for all competing algorithms and all categories. We also average over the categories and summarize the results in the Table \ref{averM}. As the Figure~\ref{mIOU} and Table~\ref{averM} suggests, proposed method consistently outperforms the competing algorithms and its variations. One interesting observation is the importance of both modalities as a result of dramatic difference between the accuracy of our method and its single modal versions.

Moreover, the difference between our method and HMM is also significant. We believe this is due to the ill-posed definition of activities in HMM since the granularity of the activity steps is subjective. On the other hand, our method starts with the well-defined definition of finding set of steps which generate the entire collection. Hence, our algorithm do not suffer from granularity problem.
\begin{table*}
\caption{\textbf{Average of $IOU_{cms}$ and $mAP_{cms}$ over recipes.}The results suggest that our algorithm outperforms all baselines. The results also suggest that both of the modalities as well as semantic representations of visual information are all required for successful parsing of video collections.}
\resizebox{2\columnwidth}{!}{%
\begin{tabular}{ccccccccc}
\toprule
 & KTS \cite{potapov2014category}    & KTS\cite{potapov2014category}     & HMM     & HMM    & Ours    & Ours     & Ours      & Our  \\
 &  w/ LLF &  w/ Sem &  w/ LLF &  w/Sem &  w/ LLF &  w/o Vis &  w/o Lang &  full \\
 \midrule
$IOU_{cms}$  & 16.80 & 28.01      & 30.84 &   37.69   &  33.16 &  36.50 & 29.91& 52.36 \\
$mAP_{cms}$  &  n/a  & n/a        & 9.35  &   32.30   &  11.33 &  30.50 &  19.50 & 44.09 \\
\bottomrule
\end{tabular}}
\normalsize
\label{averM}
\end{table*}

\paragraph{\textbf{Coherency and accuracy of activity step discovery.}}
Although $IOU_{cms}$ successfully measures the accuracy of the temporal segmentation, it can not measure the quality of discovered activities. In other words, we also need to evaluate the consistency of the activity steps detected over multiple videos. For this, we use unsupervised version of mean average precision $mAP_{cms}$. We plot the $mAP_{cms}$ values per category in Figure~\ref{mmAP} and their average over categories in Table~\ref{averM}. As the Figure~\ref{mmAP} and the Table~\ref{averM} suggests, our proposed method outperforms all competing algorithms. One interesting observation is the significant difference between semantic and low-level features. Hence, the mid-level features are key for linking multiple videos.

\paragraph{\textbf{Semantics of activity steps.}}
In order to evaluate the role of semantics, we performed a subjective analysis. We concatenated the activity step labels in the grount-truth into a label collection. Then, we ask non-expert users to choose a label for each discovered activity for each algorithm. In other words, we replaced the maximization step with subjective labels. We designed our experiments in a way that each clip received annotations from 5 different users. We randomized the ordering of videos and algorithms during the subjective evaluation. Using the labels provided by subjects, we compute the mean average precision $(mAP_{sem})$.

\begin{table}
\caption{\textbf{Semantic mean-average-precision $mAP_{sem}$.} The results suggest that our algorithm outperforms all baselines. The results also suggest that both of the modalities are required for accurate parsing for video collections.}
%{\small
\resizebox{\columnwidth}{!}{%
\begin{tabular}{ccccccc}
\toprule
            & HMM     & HMM    & Ours    & Ours     & Ours      & Our  \\
            & w/ LLF  &  w/Sem &  w/ LLF &  w/o Vis &  w/o Lang &  full \\ \midrule
$mAP_{sem}$ & 6.44   & 24.83  &     7.28 &   28.93  &  14.83    &  39.01 \\ \bottomrule
\end{tabular}}
\normalsize
\end{table}

Both $mAP_{cms}$ and $mAP_{sem}$ metrics suggest that our method consistently outperforms the competing ones. There is only one recipe in which our method is outperformed by our based line of no visual information. This is mostly because of the specific nature of the recipe \emph{How to tie a tie?}. In such videos the notion of object is not useful since all videos use a single object -tie-.
%over the entire video.
%This single object is a \emph{tie} and does not fit the assumption of a frame based on multiple visual atoms.

\paragraph{\textbf{The importance of each modality.}}
As shown in Figure~\ref{mIOU} and \ref{mmAP}, performance significantly drops when any of the modalities is ignored consistently in all categories. Hence, the joint usage is necessary. One interesting observation is the fact that using only language information performed slightly better than using only visual information. We believe this is due to the less intra-class variance in the language modality (\ie people use same words for same activities). However, it lacks many details(less complete) and more noisy than visual information. Hence these results validate the complementary nature of language and vision.

\paragraph{\textbf{Generalization to generic structured videos.}}
We experiment the applicability of our method beyond How-To videos by evaluating it on non-How-To categories. In Figure \ref{sf}, we visualize the results for the videos retrieved using the query ``Travel San Francisco". The resulting clusters follow semantically meaningful activities and landmarks and show the applicability of our method beyond How-To queries. It is interesting to note that Chinatown and Clement St ended up in the same cluster. Considering the fact that Clement St is known for its Chinese food, this suggests that the discovered clusters are semantically meaningful.

\paragraph{\textbf{Noise in the subtitles.}}
We experiment and analyze the robustness to noise in subtitles. Handling noisy subtitles is an important requirement since the scale of large-video collections makes it intractable to transcribe all instructional videos. One study suggest that it would take 374k human-year effort to transcribe all youtube videos. Hence, we expect to have combination of automatic speech recognition(ASR) generated subtitles with user uploaded ones as an input to any unsupervised parsing algorithm. 

We study the effect of noise, introduced by ASR, by evaluating our algorithm on three different video corpus. First, we only use the videos with user uploaded subtitles. Second, we only use the videos with ASR generated subtitles. Third, we use the entire dataset as union of first two. The results are summarized in Table~\ref{subt_q}. Results indicates that noise-free subtitle improves the accuracy as expected. Moreover, the difference between the results obtained with full corpus and user uploaded subtitles corpus is very small when compared with ASR only corpus. Hence, our algorithm can fuse information from noisy and noise-free examples in order to compensate for errors in the ASR.

\begin{table}[h]
\caption{\textbf{Average of $IOU_{cms}$ and $mAP_{cms}$ over recipes with and without user uploaded subtitles.} The results show that noise in the subtitles has an effect in the parsing accuracy. The results also indicate that our algorithm shows robustness to the noise since our accuracy results are comparable to the version using only user uploaded subtitles.}
\label{subt_q}
\resizebox{1\columnwidth}{!}{%
\begin{tabular}{rccc}
\toprule
& $IOU_{cms}$ & $mAP_{cms}$ & $mAP_{sem}$ \\
\midrule
ASR only & $47.61$ & $39.13$ & $33.27$ \\
User uploaded only & $54.63$ & $46.21$ & $42.32$ \\
Combination & $52.36$ & $44.09$ & $39.01$ \\
\bottomrule
\end{tabular}}
\end{table}

\section{Grounding into Robotic Instructions}
In this section, we demonstrate how we can apply our algorithm to the task of 
grounding recipe steps into robotic actions.

\begin{figure*}[ht]
\centering
  \includegraphics[width=\textwidth]{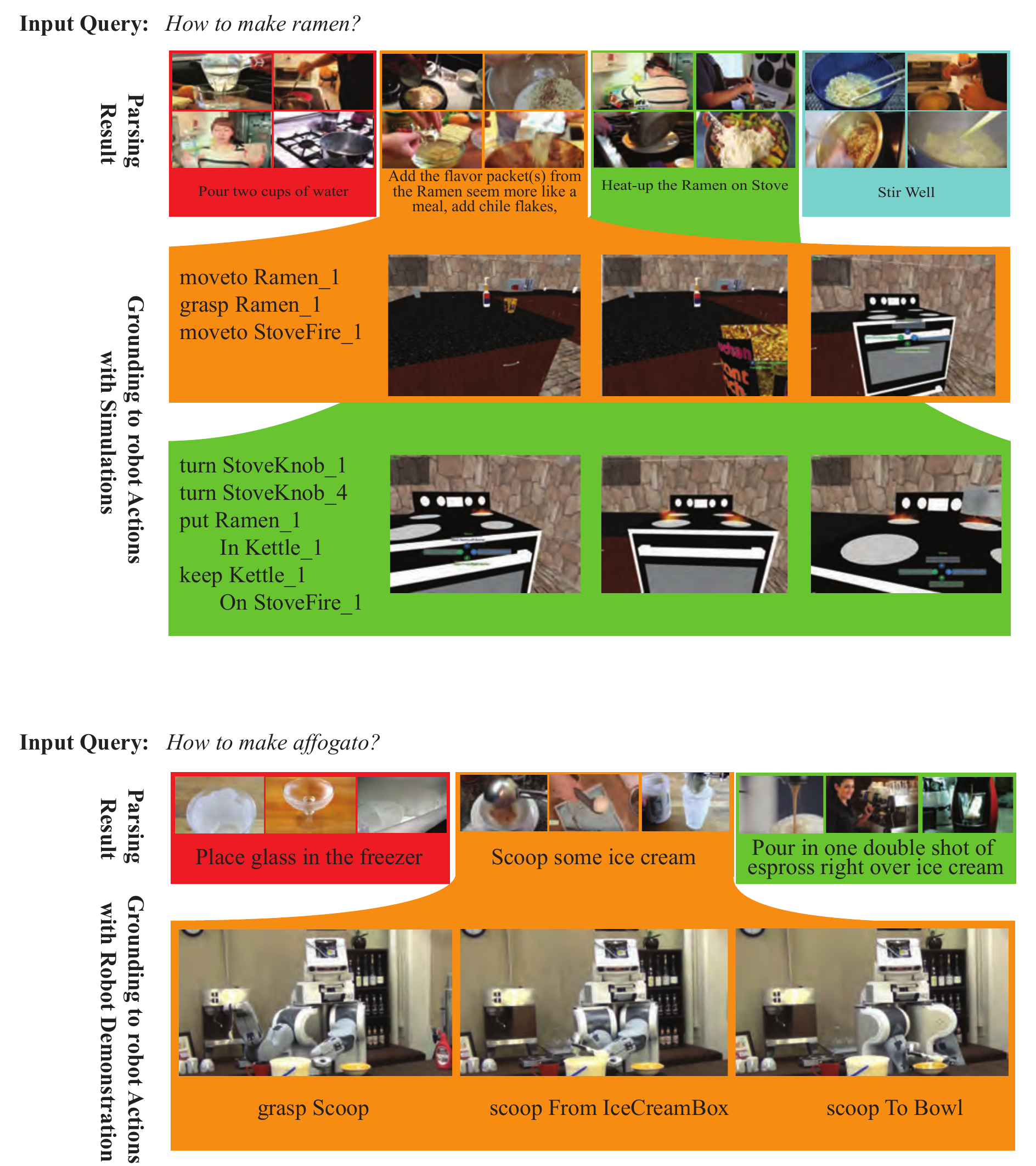} \\
  \caption{\textbf{Demonstration on robotic grounding.}  We considered two queries by the user: \emph{How to make a ramen?} and \emph{How to make an affogato?}.  Given the result by our video parsing system, we find the grounded instruction for each recipe step.  Top row shows the results as a storyline from our video parser, and the bottom row shows the robotic simulator and an actual robotic demonstration respectively. During this demonstration, we manually label each object category and fully automate rest of the task. In order to simulate/and implement the resulting steps on robots, we simply used the publicly available simulator/source code distributed by Tell Me Dave \cite{dipendra1,dipendra2}}
  \label{tmd}
\end{figure*}

One of the most important applications of our algorithms is in robotics. In future, robots will need to perform many tasks upon user's requests.  We envision that the robots can use our video parser to first download a large video collection for a task and then parse it. For example, if a user asks to the robot \emph{Please make ramen.}, the robot can download all videos returned from the query \emph{How to make a ramen.}. Robot can further parse the scene using any of the available RGB-D/RGB/Point Cloud segmentation algorithms \cite{hema, xiaofeng, iro}. Robot can use the resulting segmentation in order to find the most similar recipe simply using the object categories that we output.

In order to demonstrate this application, we use a state of the art language grounding algorithm \cite{dipendra1, dipendra2} which  converts the generated descriptions into robot actions based on the environment. Tell Me Dave algorithm of Misra et al \cite{dipendra1,dipendra2} uses a semantic simulator which encodes the common sense knowledge about the physical world. It takes the tuple of language, instructions and the environment as an input and outputs a series of robot actions to perform the task. In order to learn the transformation, it uses a large-scale game log of language instruction, environment and robot action tuples, and models them in terms of a graphical model. The environment is defined in terms of objects and their 3D positions, language is series of free-form English sentences describing each step and actions are low-level robot commands.

In our experimental setup, we choose two basic activites that Tell Me Dave can simulate; namely, \emph{How to make a ramen?} and \emph{How to make an affogato?}. We also chose a random environment for each query from Tell Me Dave environment dataset. We directly feed aforementioned how-to queries into YouTube and parse the resulting video collections. The resulting storylines are visualized in Figure~\ref{tmd}.
 
In order to complete the loop until low-level robot commands, we then manually labelled the object categories our algorithm discovered. For example, if the category we discovered is mostly images of eggs, we labelled this category as \emph{egg}. \footnote{This step can be automated using any object recognition algorithm \cite{imagenet}.} Using these labels, our algorithm chose the video whose objects is a subset to the environment our robot lives in to make sure all objects of the recipe are accessible by the robot. Finally, we feed the environment and generated captions into the Tell Me Dave algorithm to obtain the physical plan robot needs to execute to perform each of the activity. We visualize each plan and the simulation in the Figure~\ref{tmd}.

Our results are shown in the Figure~\ref{tmd}, demonstrating that our approach can be used for robotics applications with limited supervision. There were some errors in translation of video storyline steps to actual grounded steps. Example errors include turning on both of the knobs of the stove other than the single one. However, the resulting plans were still feasible in a way they can accomplish the required high-level task. 

While a larger analysis and robotic experiments are outside the scope of this paper, with this demonstration we believe that our proposed method shows a feasible direction for robotics.

% !TEX root = main.tex

\section{Conclusions}
%Discuss which recipes worked and why. Discuss the importance of semantic representation, scaling features and multi-modality.
In this paper, we captured the underlying structure of human communication by jointly considering visual and language cues. We experimentally validate that given a large-video collection having subtitles, it is possible to discover activities without any supervision over activities or objects. Experimental evaluation also suggests the available noisy and incomplete information is powerful enough to not only discover activities but also describe them. We also demonstrated that the resulting discovered recipes are useful in robotics scenarios.

So; ``is it possible to understand large-video collections without any supervision?''. Given video and speech information, the storylines we generate successfully summarize the large video collection. This compact representation of an hundreds of videos is expected to be useful in designing user interfaces which users interact with instructional videos. We believe this is an important step in the direction of future video webpages. Yet another very important question is ``can machines understand large-video collections?''. Clearly, we needed a small amount of manual information and even used a method which is trained with human supervision in our robotic demonstration. Hence, it is too early to claim a success for machines watching large-collection of videos. On the other hand, the results are very promising and we believe algorithms which can convert a free-form input query into robot trajectories are a possible in near future. We also believe our algorithm is an important step in this ambitious target.

\appendix
\section{Generation of Language Description}
\label{langgen}
In this section, we explain how we generated the text description for the activity steps we discovered. We included these descriptions in Figure~\ref{recipe:overall},~\ref{sf}\&\ref{tmd} as well as in the supplementary videos.

In order to generate the descriptions, we simply used a Markov text generator. We collected all subtitles of all videos we included in our dataset. After combining them, we trained a $3^{rd}$ order Markov model by using the subtitles we downloaded. Main purpose of this training is learning the context dependent language model. Although this step can be accomplished by various of methods in the NLP literature, we choose Markov language model because of its simplicity. Indeed, this model is learned purely for visualization purposes and neither the activity step discovery nor the parsing algorithm uses this model.

After the model is learned, we need to generate a text description for each discovered activity. Since each discovered activity is represented as Bernoulli random variable, we have likelihood for each language atom. Our description generation strategy is sampling a large collection of descriptions and ranking them for their closeness to the discovered activities. We compute this closeness with the parameters of the Bernoulli random variable. Formally, given large-set of sampled descriptions $\{S_i\}_{i \in [1,K]}$, we rank them using the weights of the Bernoulli random variable as;
\[
r_i = \frac{\sum_{j} \left[S_i^j = w^j\right]\theta^j}{\sum_{j} \left[S_i^j = w^j\right]1}
\]
Here, $\left[\cdot\right]$ is an indicator function, $S_i^j$ is the $j^{th}$ word of $i^{th}$ description, $w^j$ is the $j^{th}$ word and $\theta^j$ is the $j^{th}$ entry of the activity description. We simply choose the description having largest rank.

\clearpage

  \section*{Acknowledgments}

The authors would like to thank Jay Hack for developing ModalDB and Bart Selman and Ashesh Jain for useful comments and discussions.

\bibliographystyle{spbasic}

\begin{thebibliography}{69}
\providecommand{\natexlab}[1]{#1}
\providecommand{\url}[1]{{#1}}
\providecommand{\urlprefix}{URL }
\expandafter\ifx\csname urlstyle\endcsname\relax
  \providecommand{\doi}[1]{DOI~\discretionary{}{}{}#1}\else
  \providecommand{\doi}{DOI~\discretionary{}{}{}\begingroup
  \urlstyle{rm}\Url}\fi
\providecommand{\eprint}[2][]{\url{#2}}

\bibitem[{wik(2015)}]{wikiHow}
 (2015) Wikihow-how to do anything. \url{http://www.wikihow.com}

\bibitem[{Armeni et~al(2016)Armeni, Sener, Zamir, and Savarese}]{iro}
Armeni I, Sener O, Zamir A, Savarese S (2016) 3d semantic parsing of
  large-scale indoor spaces. In: CVPR

\bibitem[{Barbu et~al(2012)Barbu, Bridge, Burchill, Coroian, Dickinson, Fidler,
  Michaux, Mussman, Narayanaswamy, Salvi et~al}]{barbu2012video}
Barbu A, Bridge A, Burchill Z, Coroian D, Dickinson S, Fidler S, Michaux A,
  Mussman S, Narayanaswamy S, Salvi D, et~al (2012) Video in sentences out.
  arXiv preprint arXiv:12042742

\bibitem[{Barnard et~al(2003)Barnard, Duygulu, Forsyth, De~Freitas, Blei, and
  Jordan}]{matching}
Barnard K, Duygulu P, Forsyth D, De~Freitas N, Blei DM, Jordan MI (2003)
  Matching words and pictures. JMLR 3:1107--1135

\bibitem[{Beetz et~al(2011)Beetz, Klank, Kresse, Maldonado, Mosenlechner,
  Pangercic, Ruhr, and Tenorth}]{beetz}
Beetz M, Klank U, Kresse I, Maldonado A, Mosenlechner L, Pangercic D, Ruhr T,
  Tenorth M (2011) Robotic roommates making pancakes. In: Humanoids

\bibitem[{Bojanowski et~al(2014)Bojanowski, Lajugie, Bach, Laptev, Ponce,
  Schmid, and Sivic}]{bojanowski14_eccv}
Bojanowski P, Lajugie R, Bach F, Laptev I, Ponce J, Schmid C, Sivic J (2014)
  Weakly supervised action labeling in videos under ordering constraints. In:
  ECCV

\bibitem[{Bollini et~al(2011)Bollini, Barry, and Rus}]{cookie}
Bollini M, Barry J, Rus D (2011) Bakebot: Baking cookies with the pr2. In: The
  PR2 Workshop, IROS

\bibitem[{Carreira and Sminchisescu(2010)}]{cpmc}
Carreira J, Sminchisescu C (2010) Constrained parametric min-cuts for automatic
  object segmentation. In: CVPR

\bibitem[{Das et~al(2013)Das, Xu, Doell, and Corso}]{das2013thousand}
Das P, Xu C, Doell RF, Corso JJ (2013) A thousand frames in just a few words:
  Lingual description of videos through latent topics and sparse object
  stitching. In: CVPR

\bibitem[{Duchenne et~al(2009)Duchenne, Laptev, Sivic, Bash, and
  Ponce}]{duchenne09_iccv}
Duchenne O, Laptev I, Sivic J, Bash F, Ponce J (2009) Automatic annotation of
  human actions in video. In: ICCV

\bibitem[{Efros et~al(2003)Efros, Berg, Mori, and Malik}]{efros03_iccv}
Efros AA, Berg AC, Mori G, Malik J (2003) Recognizing action at a distance. In:
  ICCV

\bibitem[{Farhadi et~al(2010)Farhadi, Hejrati, Sadeghi, Young, Rashtchian,
  Hockenmaier, and Forsyth}]{farhadi2010every}
Farhadi A, Hejrati M, Sadeghi MA, Young P, Rashtchian C, Hockenmaier J, Forsyth
  D (2010) Every picture tells a story: Generating sentences from images

\bibitem[{Fidler et~al(2013)Fidler, Sharma, and Urtasun}]{fidler2013sentence}
Fidler S, Sharma A, Urtasun R (2013) A sentence is worth a thousand pixels. In:
  CVPR, IEEE

\bibitem[{Fox et~al(2014)Fox, Hughes, Sudderth, and Jordan}]{foxBPHMM}
Fox E, Hughes M, Sudderth E, Jordan M (2014) Joint modeling of multiple related
  time series via the beta process with application to motion capture
  segmentation. Annals of Applied Statistics 8(3):1281--1313

\bibitem[{Google(2016)}]{google_trends}
Google (2016) Google trends. \url{https://www.google.com/trends/}, accessed:
  2016-04-23

\bibitem[{Grabler et~al(2009)Grabler, Agrawala, Li, Dontcheva, and
  Igarashi}]{photoshop}
Grabler F, Agrawala M, Li W, Dontcheva M, Igarashi T (2009) Generating photo
  manipulation tutorials by demonstration. TOG 28(3):66

\bibitem[{Griffiths and Ghahramani(2005)}]{ibp}
Griffiths T, Ghahramani Z (2005) Infinite latent feature models and the indian
  buffet process. Gatsby Unit

\bibitem[{Gupta et~al(2009)Gupta, Srinivasan, Shi, and
  Davis}]{gupta2009understanding}
Gupta A, Srinivasan P, Shi J, Davis LS (2009) Understanding videos,
  constructing plots learning a visually grounded storyline model from
  annotated videos. In: CVPR

\bibitem[{Hoai et~al(2011)Hoai, Lan, and {De la Torre}}]{hoai11_cvpr}
Hoai M, Lan ZZ, {De la Torre} F (2011) Joint segmentation and classification of
  human actions in video. In: CVPR

\bibitem[{Hughes and Sudderth(2012)}]{npActivity}
Hughes MC, Sudderth EB (2012) Nonparametric discovery of activity patterns from
  video collections. In: CVPR Workshop, pp 25--32

\bibitem[{Jain et~al(2013)Jain, Jegou, and Bouthemy}]{jain13_cvpr}
Jain M, Jegou H, Bouthemy P (2013) Better exploiting motion for better action
  recognition. In: CVPR

\bibitem[{Jain et~al(2014)Jain, van Gemert, and Snoek}]{jainuniversity}
Jain M, van Gemert J, Snoek CG (2014) University of amsterdam at thumos
  challenge 2014. ECCV International Workshop and Competition on Action
  Recognition with a Large Number of Classes

\bibitem[{Jiang et~al(2014)Jiang, Liu, Roshan~Zamir, Toderici, Laptev, Shah,
  and Sukthankar}]{THUMOS14}
Jiang YG, Liu J, Roshan~Zamir A, Toderici G, Laptev I, Shah M, Sukthankar R
  (2014) {THUMOS} challenge: Action recognition with a large number of classes.
  \url{http://crcv.ucf.edu/THUMOS14/}

\bibitem[{{Karpathy} and {Fei-Fei}(2014)}]{deepAlignment}
{Karpathy} A, {Fei-Fei} L (2014) {Deep Visual-Semantic Alignments for
  Generating Image Descriptions}. ArXiv e-prints \eprint{1412.2306}

\bibitem[{Khosla et~al(2013)Khosla, Hamid, Lin, and
  Sundaresan}]{khosla2013large}
Khosla A, Hamid R, Lin CJ, Sundaresan N (2013) Large-scale video summarization
  using web-image priors. In: CVPR

\bibitem[{Kim and Xing(2014)}]{storyGraph}
Kim G, Xing EP (2014) Reconstructing storyline graphs for image recommendation
  from web community photos. In: CVPR

\bibitem[{Kim et~al(2014)Kim, Sigal, and Xing}]{kim2014joint}
Kim G, Sigal L, Xing EP (2014) Joint summarization of large-scale collections
  of web images and videos for storyline reconstruction. In: CVPR

\bibitem[{Kiros et~al(2014)Kiros, Salakhutdinov, and
  Zemel}]{kiros2014multimodal}
Kiros R, Salakhutdinov R, Zemel R (2014) Multimodal neural language models. In:
  ICML

\bibitem[{Kong et~al(2014)Kong, Lin, Bansal, Urtasun, and Fidler}]{kong2014you}
Kong C, Lin D, Bansal M, Urtasun R, Fidler S (2014) What are you talking about?
  text-to-image coreference. In: CVPR

\bibitem[{Koppula et~al(2011)Koppula, Anand, Joachims, and Saxena}]{hema}
Koppula HS, Anand A, Joachims T, Saxena A (2011) Semantic labeling of 3d point
  clouds for indoor scenes. In: Shawe-Taylor J, Zemel RS, Bartlett PL, Pereira
  F, Weinberger KQ (eds) Advances in Neural Information Processing Systems 24,
  Curran Associates, Inc., pp 244--252,
  \urlprefix\url{http://papers.nips.cc/paper/4226-semantic-labeling-of-3d-point-clouds-for-indoor-scenes.pdf}

\bibitem[{Krizhevsky et~al(2012)Krizhevsky, Sutskever, and Hinton}]{alexnet}
Krizhevsky A, Sutskever I, Hinton GE (2012) Imagenet classification with deep
  convolutional neural networks. In: NIPS

\bibitem[{Kuehne et~al(2011)Kuehne, Jhuang, Garrote, Poggio, and
  Serre}]{kuehne2011hmdb}
Kuehne H, Jhuang H, Garrote E, Poggio T, Serre T (2011) Hmdb: a large video
  database for human motion recognition. In: ICCV

\bibitem[{Lan et~al(2014{\natexlab{a}})Lan, Chen, Deng, Zhou, and
  Mori}]{lan14_vs}
Lan T, Chen L, Deng Z, Zhou GT, Mori G (2014{\natexlab{a}}) Learning action
  primitives for multi-level video event understanding. In: Workshop on Visual
  Surveillance and Re-Identification

\bibitem[{Lan et~al(2014{\natexlab{b}})Lan, Chen, and Savarese}]{lan14_eccv}
Lan T, Chen TC, Savarese S (2014{\natexlab{b}}) A hierarchical representation
  for future action prediction. In: ECCV

\bibitem[{Laptev and P\'{e}rez(2007)}]{laptev07_iccv}
Laptev I, P\'{e}rez P (2007) Retrieving actions in movies. In: ICCV

\bibitem[{Laptev et~al(2008)Laptev, Marszalek, Schmid, and
  Rozenfeld}]{laptev08_cvpr}
Laptev I, Marszalek M, Schmid C, Rozenfeld B (2008) Learning realistic human
  actions from movies. In: CVPR

\bibitem[{Lee et~al(2011)Lee, Kim, and Grauman}]{keysegments}
Lee YJ, Kim J, Grauman K (2011) Key-segments for video object segmentation. In:
  ICCV

\bibitem[{Lee et~al(2012)Lee, Ghosh, and Grauman}]{lee2012discovering}
Lee YJ, Ghosh J, Grauman K (2012) Discovering important people and objects for
  egocentric video summarization. In: CVPR

\bibitem[{Liao(2005)}]{liao05}
Liao TW (2005) Clustering of time series data—a survey. Pattern recognition
  38(11):1857--1874

\bibitem[{Lu and Grauman(2013)}]{lu2013story}
Lu Z, Grauman K (2013) Story-driven summarization for egocentric video. In:
  CVPR

\bibitem[{Malmaud et~al(2014)Malmaud, Wagner, Chang, and
  Murphy}]{cookingSemantics}
Malmaud J, Wagner EJ, Chang N, Murphy K (2014) Cooking with semantics. ACL

\bibitem[{{Malmaud} et~al(2015){Malmaud}, {Huang}, {Rathod}, {Johnston},
  {Rabinovich}, and {Murphy}}]{alignment}
{Malmaud} J, {Huang} J, {Rathod} V, {Johnston} N, {Rabinovich} A, {Murphy} K
  (2015) {What's Cookin'? Interpreting Cooking Videos using Text, Speech and
  Vision}. ArXiv e-prints \eprint{1503.01558}

\bibitem[{Misra et~al(2014)Misra, Sung, Lee, and Saxena}]{dipendra1}
Misra DK, Sung J, Lee K, Saxena A (2014) Tell me dave: Context-sensitive
  grounding of natural language to mobile manipulation instructions. In: In
  RSS, Citeseer

\bibitem[{Misra et~al(2015)Misra, Tao, Liang, and Saxena}]{dipendra2}
Misra DK, Tao K, Liang P, Saxena A (2015) Environment-driven lexicon induction
  for high-level instructions. ACL

\bibitem[{Motwani and Mooney(2012)}]{motwani2012improving}
Motwani TS, Mooney RJ (2012) Improving video activity recognition using object
  recognition and text mining. In: ECAI

\bibitem[{Niebles et~al(2010)Niebles, Chen, and Fei-Fei}]{niebles10_eccv}
Niebles JC, Chen CW, Fei-Fei L (2010) Modeling temporal structure of
  decomposable motion segments for activity classification. In: ECCV

\bibitem[{Olson et~al(2005)Olson, Walter, Teller, and Leonard}]{scgp}
Olson E, Walter M, Teller SJ, Leonard JJ (2005) Single-cluster spectral graph
  partitioning for robotics applications. In: RSS

\bibitem[{Oneata et~al(2014)Oneata, Verbeek, and Schmid}]{oneata2014lear}
Oneata D, Verbeek J, Schmid C (2014) The lear submission at thumos 2014. ECCV
  International Workshop and Competition on Action Recognition with a Large
  Number of Classes

\bibitem[{Ordonez et~al(2011)Ordonez, Kulkarni, and Berg}]{ordonez2011im2text}
Ordonez V, Kulkarni G, Berg TL (2011) Im2text: Describing images using 1
  million captioned photographs. In: NIPS

\bibitem[{Perona and Freeman(1998)}]{scgp_eigen}
Perona P, Freeman W (1998) A factorization approach to grouping. In: ECCV

\bibitem[{Pirsiavash and Ramanan(2014)}]{pirsiavash14_cvpr}
Pirsiavash H, Ramanan D (2014) Parsing videos of actions with segmental
  grammars. In: CVPR

\bibitem[{Potapov et~al(2014)Potapov, Douze, Harchaoui, and
  Schmid}]{potapov2014category}
Potapov D, Douze M, Harchaoui Z, Schmid C (2014) Category-specific video
  summarization. In: ECCV

\bibitem[{Rabiner(1989)}]{rabiner}
Rabiner LR (1989) A tutorial on hidden markov models and selected applications
  in speech recognition. In: PROCEEDINGS OF THE IEEE, pp 257--286

\bibitem[{Ren et~al(2012)Ren, Bo, and Fox}]{xiaofeng}
Ren X, Bo L, Fox D (2012) Rgb-(d) scene labeling: Features and algorithms. In:
  CVPR

\bibitem[{Rui et~al(2000)Rui, Gupta, and Acero}]{rui2000automatically}
Rui Y, Gupta A, Acero A (2000) Automatically extracting highlights for tv
  baseball programs. In: ACM MM

\bibitem[{Russakovsky et~al(2015)Russakovsky, Deng, Su, Krause, Satheesh, Ma,
  Huang, Karpathy, Khosla, Bernstein et~al}]{imagenet}
Russakovsky O, Deng J, Su H, Krause J, Satheesh S, Ma S, Huang Z, Karpathy A,
  Khosla A, Bernstein M, et~al (2015) Imagenet large scale visual recognition
  challenge. International Journal of Computer Vision 115(3):211--252

\bibitem[{Ryoo and Aggarwal(2009)}]{ryoo09_iccv}
Ryoo M, Aggarwal J (2009) Spatio-temporal relationship match: Video structure
  comparison for recognition of complex human activities. In: ICCV

\bibitem[{Shannon(2001)}]{languageModel}
Shannon CE (2001) A mathematical theory of communication. ACM SIGMOBILE Mobile
  Computing and Communications Review 5(1):3--55

\bibitem[{Socher and Fei-Fei(2010)}]{connecting}
Socher R, Fei-Fei L (2010) Connecting modalities: Semi-supervised segmentation
  and annotation of images using unaligned text corpora. In: CVPR, pp 966--973

\bibitem[{Socher et~al(2014)Socher, Karpathy, Le, Manning, and
  Ng}]{socher2014grounded}
Socher R, Karpathy A, Le QV, Manning CD, Ng AY (2014) Grounded compositional
  semantics for finding and describing images with sentences. TACL 2:207--218

\bibitem[{Soomro et~al(2012)Soomro, Roshan~Zamir, and Shah}]{UCF101}
Soomro K, Roshan~Zamir A, Shah M (2012) {UCF101}: A dataset of 101 human
  actions classes from videos in the wild. In: CRCV-TR-12-01

\bibitem[{Sun and Nevatia(2014)}]{sun2014discover}
Sun C, Nevatia R (2014) Discover: Discovering important segments for
  classification of video events and recounting. In: CVPR

\bibitem[{Tenorth et~al(2010)Tenorth, Nyga, and Beetz}]{logicRecipe}
Tenorth M, Nyga D, Beetz M (2010) Understanding and executing instructions for
  everyday manipulation tasks from the world wide web. In: ICRA

\bibitem[{Truong and Venkatesh(2007)}]{vidAbstraction}
Truong BT, Venkatesh S (2007) Video abstraction: A systematic review and
  classification. ACM TOMM 3(1):3

\bibitem[{Wikipedia(2004)}]{wiki}
Wikipedia (2004) {W}ikipedia{,} the free encyclopedia.
  \urlprefix\url{http://en.wikipedia.org/}, [Online; accessed 22-April-2016]

\bibitem[{Yao and Fei-Fei(2010)}]{yao10b_cvpr}
Yao B, Fei-Fei L (2010) Modeling mutual context of object and human pose in
  human-object interaction activities. In: CVPR

\bibitem[{Yu and Siskind(2013)}]{yu2013grounded}
Yu H, Siskind JM (2013) Grounded language learning from video described with
  sentences. In: ACL

\bibitem[{Zitnick and Parikh(2013)}]{zitnick2013bringing}
Zitnick CL, Parikh D (2013) Bringing semantics into focus using visual
  abstraction. In: CVPR

\bibitem[{Zitnick et~al(2013)Zitnick, Parikh, and
  Vanderwende}]{zitnick2013learning}
Zitnick CL, Parikh D, Vanderwende L (2013) Learning the visual interpretation
  of sentences. In: CVPR

\end{thebibliography}

\end{document}